\documentclass[acmsmall]{acmart}

\setcopyright{licensedcagov}
\copyrightyear{2025}
\acmYear{2025}
\acmDOI{10.1145/3721983}

\usepackage{amsmath,amsfonts}
\usepackage{mathtools}
\usepackage{graphicx}
\usepackage{xcolor}
\usepackage{textcomp}

\usepackage{float}
\usepackage{subfig}         

\usepackage{array}
\usepackage{tabularx}
\usepackage{multirow}
\usepackage{longtable}
\usepackage{adjustbox}
\usepackage{booktabs}       

\usepackage{placeins}
\usepackage{dblfloatfix}
\usepackage{afterpage}

\usepackage{caption}
\usepackage{cleveref}

\usepackage{optidef}
\usepackage{svg}
\usepackage[titlenumbered,ruled,linesnumbered]{algorithm2e}

\newtheorem{definition}{Definition}

\AtBeginDocument{%
  \providecommand\BibTeX{{%
    \normalfont B\kern-0.5em{\scshape i\kern-0.25em b}\kern-0.8em\TeX}}}

\acmJournal{TKDD}
\acmMonth{3}

\begin{document}

\title{SED2AM: Solving Multi-Trip Time-Dependent Vehicle Routing Problem using Deep Reinforcement Learning}

\author{Arash Mozhdehi}
\email{arash.mozhdehi@ucalgary.ca}
\orcid{0000-0002-9938-3560}
\affiliation{
  \institution{University of Calgary}
  \streetaddress{2500 University Dr. NW}
  \city{Calgary}
  \state{Alberta}
  \country{Canada}
  \postcode{T2N 1N4}
}

\author{Yunli Wang}
\affiliation{
  \institution{National Research Council Canada}
  \city{Ottawa}
  \country{Canada}}
\email{yunli.wang@nrc-cnrc.gc.ca}
\orcid{0000-0002-2320-954X}

\author{Sun Sun}
\affiliation{
  \institution{National Research Council Canada}
  \city{Waterloo}
  \country{Canada}}
\email{sun.sun@nrc-cnrc.gc.ca}
\orcid{0000-0001-7870-9448}

\author{Xin Wang}
\affiliation{
  \institution{University of Calgary}
  \streetaddress{2500 University Dr. NW}
  \city{Calgary}
  \state{Alberta}
  \country{Canada}
  \postcode{T2N 1N4}}
\email{xcwang@ucalgary.ca}
\orcid{0000-0003-3569-2126}

\renewcommand{\shortauthors}{Mozhdehi et al.}

\begin{abstract}
Deep reinforcement learning (DRL)-based frameworks, featuring Transformer-style policy networks, have demonstrated their efficacy across various vehicle routing problem (VRP) variants. However, the application of these methods to the multi-trip time-dependent vehicle routing problem (MTTDVRP) with maximum working hours constraints—a pivotal element of urban logistics—remains largely unexplored. This paper introduces a DRL-based method called the Simultaneous Encoder and Dual Decoder Attention Model (SED2AM), tailored for the MTTDVRP with maximum working hours constraints. The proposed method introduces a temporal locality inductive bias to the encoding module of the policy networks, enabling it to effectively account for the time-dependency in travel distance/time. The decoding module of SED2AM includes a vehicle selection decoder that selects a vehicle from the fleet, effectively associating trips with vehicles for functional multi-trip routing. Additionally, this decoding module is equipped with a trip construction decoder leveraged for constructing trips for the vehicles. This policy model is equipped with two classes of state representations, fleet state and routing state, providing the information needed for effective route construction in the presence of maximum working hours constraints. Experimental results using real-world datasets from two major Canadian cities not only show that SED2AM outperforms the current state-of-the-art DRL-based and metaheuristic-based baselines but also demonstrate its generalizability to solve larger-scale problems.
\end{abstract}



\received{21 March 2024}
\received[revised]{16 December 2024}
\received[accepted]{27 February 2025}
\maketitle

\section{Introduction}
The multi-trip time-dependent vehicle routing problem with the maximum working hours constraints (MTTDVRP) is a practical variant of the traditional vehicle routing problem (VRP) tailored to meet the complex demands of urban logistics. MTTDVRP seeks to determine a set of trips, each starting and ending at the depot, as illustrated in Figure \ref{fig:mttdvrp_example}, for a fleet of vehicles with the goal of minimizing total travel time under traffic conditions that vary throughout the day. This optimization problem allows each vehicle to perform multiple trips within a single working day, provided that the cumulative travel time across trips for each vehicle does not exceed the maximum allowable working hours set for the driver, ensuring compliance with traffic regulations \cite{legislative}. Figure \ref{fig:mttdvrp_example} illustrates an example of MTTDVRP with the maximum working hours constraints, showing multiple trips carried out by two different vehicles within a fleet, while adhering to the maximum working hours for each vehicle and the maximum capacity constraint for each individual trip.

\begin{center}
    \includegraphics[width=5.5 in]{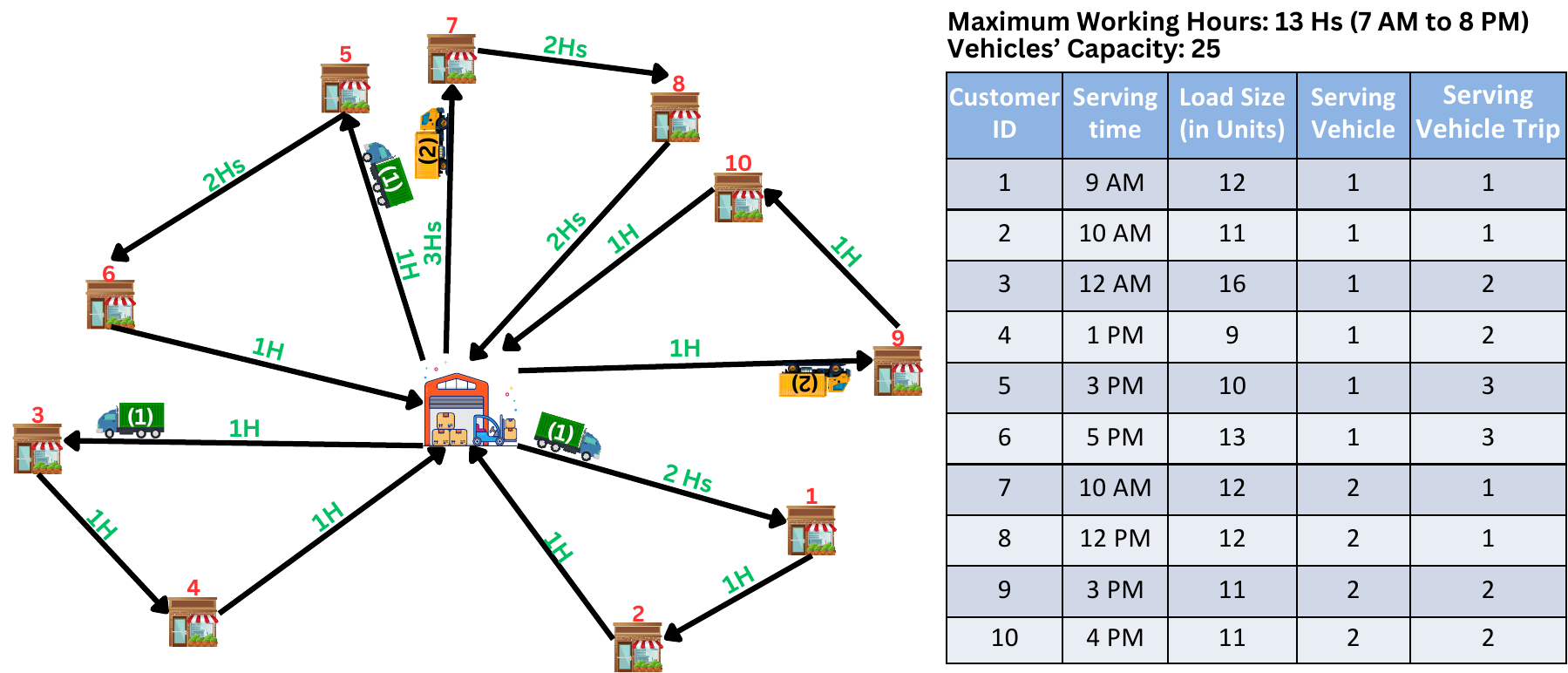}
    \captionof{figure}{Example of MTTDVRP with maximum working hours constraints. This illustration depicts multiple trips conducted by two vehicles within a fleet, adhering to a maximum working hours limit of 13 hours (7 AM to 8 PM) and a vehicle capacity of 25 units per trip. The red numbers above each customer location represent the customer number, while the green labels indicate travel times between each origin and destination based on traffic conditions. The table summarizes the time each customer is served, the customer's demand, and which vehicle on which trip served the customer.}
    \label{fig:mttdvrp_example}
\end{center}

VRPs, as NP-hard combinatorial optimization problems, were traditionally solved using exact and heuristic methods. Although exact methods, such as branch-and-bound, provide optimal solutions, they struggle with exponential computational complexity, making them impractical for real-world problems that involve serving a large number of customers. On the other hand, heuristic solvers, such as the Clarke and Wright saving algorithm and the genetic algorithm, deliver sub-optimal solutions within a reasonable computational time. However, these solvers suffer from limitations that have been widely discussed in the literature \cite{burke2013hyper, bello2016neural, li2020deep}, such as a lack of generalizability and reliance on human-crafted heuristics. Recently, deep reinforcement learning (DRL)-based methods \cite{bello2016neural, nazari2018reinforcement} have been employed to overcome the shortcomings of the traditional methods. Among these DRL-based methods, those employing Transformer-style policy networks \cite{kool2018attention, li2020deep, xu2021reinforcement, li2021heterogeneous, zong2022mapdp, zhang2022meta, chen2023reinforcement, zong2024reinforcement} showed promising results. Despite the success of this class of DRL-based methods in solving various VRPs, they are ineffective in solving MTTDVRP with maximum working hours constraints due to the following shortcomings:

\begin{enumerate}
\item \textbf{Intrinsic inability to capture the dynamicity in travel times:} 
For DRL-based approaches, routing effectiveness critically depends on accurately modeling the dynamic relationship between customers and the depot \cite{xu2021reinforcement, li2024solving, wang2024deep}. A key component of this relationship is the travel time and distance between locations, which directly impact route optimization \cite{lei2022solve, duan2020efficiently}. However, existing methods fall short in addressing the variability of travel time and distance due to fluctuating traffic conditions over time. This limitation arises because these methods often employ encoders restricted to embedding static, one-dimensional features, thereby neglecting the critical time-dependent dimension. However, in reality, travel time and even distance (such as road closure) between locations are not fixed and are influenced by fluctuating traffic and road conditions (e.g., road closures). Figure \ref{fig:td_dist} illustrates how the same locations can yield different optimal routes under changing traffic conditions.

\item \textbf{Failure to directly associate the constructed trips to the corresponding vehicle:} 
For multi-trip routing, the cost of each trip, starting and ending at the depot, can vary significantly depending on the starting time due to fluctuating traffic conditions throughout the day. Consequently, trips with identical sequences of customer may incur different costs based on their initiation times, influenced by the vehicle’s previous trips. However, the existing methods fail to associate the trips undertaken to the respective vehicles, planning routes independently instead. These DRL-based approaches do not allow the routes to be constructed based on the actual timing of trips, thereby rendering these methods ineffective for multi-trip, time-dependent routing.

\item \textbf{Insufficiency of vehicle-specific state information:} When a fleet is performing multi-trip routing under maximum working hours constraints, allowing the model to select vehicles based on remaining working hours as state information can aid the policy model in rational decision-making. In fact, with this routing problem variant, vehicles are constrained not only by their capacity but also by their limited maximum operational time. The existing methods do not utilize remaining working hours in the decision-making process as a critical vehicle-specific state information, rendering them ineffective for multi-trip routing when maximum working hours constraints are in place. Additionally, for time-dependent routing, the ability to effectively construct routes is closely tied to the traffic conditions that the vehicle will encounter during the trips, which, in turn, depend on the location and starting time of the trips. Capturing this information in the state is critical for time-dependent routing; however, existing methods fail to exploit this information in their state representation.
 
\begin{figure}[h!]
\centering
    \subfloat[Starting at 7 AM]{
        \label{diff_dist_map1}
        \includegraphics[width=2.2in]{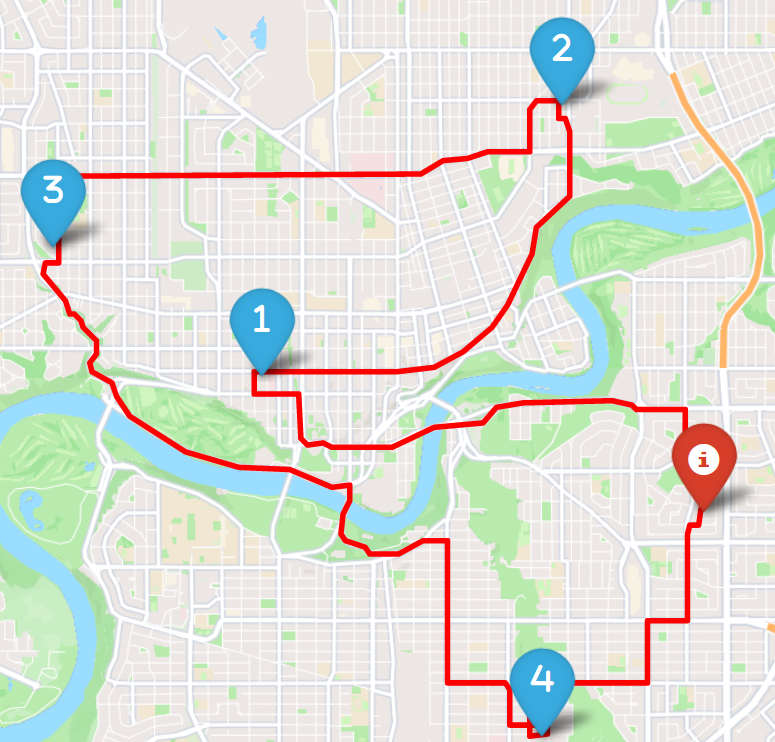}
    }
    \subfloat[Starting at 4 PM]{
        \label{diff_dist_map2}
        \includegraphics[width=2.2in]{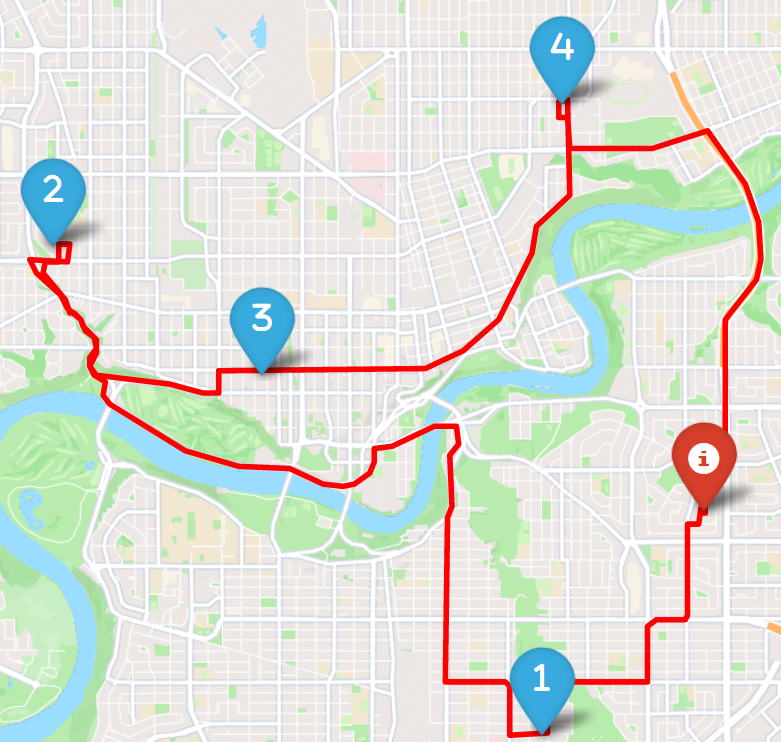}
    }
    \caption{Cases where the same locations resulted in different optimal routes due to varying traffic conditions at different times of the day, 7 AM vs. 4 PM. Blue pin-drops indicate customers' locations, with numbers showing the visiting order, and the red pin-drop represents the depot's location.}
\label{fig:td_dist}
\end{figure}

\end{enumerate}

In this paper, to effectively solve the MTTDVRP with a maximum working hours constraints, we propose a DRL-based approach named the Simultaneous Encoder and Dual Decoder Attention Model (SED2AM). The major contributions of this paper can be summarized as follows:
\begin{itemize}
    \item To effectively capture the time dependency of travel distances and times, we introduce a temporal locality inductive bias to the Transformer-based policy network. This approach is based on the rationale that selecting the next destination for each vehicle is primarily influenced by the immediate and relevant traffic conditions encountered during travel, rather than by traffic conditions at other times. Based on this inductive bias, we design a Transformer-style encoder, referred to as the simultaneous encoder, to capture the relationships between locations in alignment with the dynamic nature of real-world traffic. This encoder co-embeds the features of each node—representing either the depot or a customer—along with the time-dependent travel times between pairs of locations as edge features. Considering the introduced inductive bias, this co-embedding operation is performed separately for different periods of the day, ensuring the effective capture of time dependency while minimizing computational complexity.    
  
    \item To enable the model to effectively associate routes constructed by each vehicle, a dual-decoder attention model is proposed. This model includes a vehicle selection decoder that chooses a vehicle from the fleet to expand its current route, allowing each vehicle to track its own routes. This decoder employs a simple yet effective network for fast and efficient multi-trip routing. Following vehicle selection, the trip construction decoder identifies the next location—either a depot or customer—for the vehicle to visit, using an attention-based mechanism. Additionally, we introduced a state representation comprising the fleet state and routing state, providing the policy model with sufficient vehicle-specific information to make informed decisions under maximum working hours constraints in time-dependent multi-trip vehicle routing scenarios.
    \item  We evaluate SED2AM's performance on problem sets derived from Edmonton and Calgary, using real traffic data to estimate travel times. Experimental results show that SED2AM outperforms heuristic- and state-of-the-art DRL-based approaches in total travel time, while achieving faster runtimes than traditional methods. Additionally, SED2AM demonstrates strong generalizability to larger problem instances.
\end{itemize}

The remainder of this paper is organized as follows: Section \ref{section:Related_Works} briefly reviews related work. In Section \ref{section:preliminaries}, we formally describe and define the MTTDVRP with the maximum working hours constraints. We elaborate on SED2AM in Section \ref{section:method}. In Section \ref{section:experiments}, the computational experiments and analysis are presented. Finally, Section \ref{section:conclusion} concludes this paper.

\section{Related Works}
\label{section:Related_Works}

In this section, we briefly review the traditional methods for solving VRP variants, including multi-trip time-dependent VRP (MTTDVRP), as well as approaches in the DRL literature for solving general VRPs.

MTTDVRP is a variant of VRP that has rarely been studied, but given its application for real-world problems, merits more investigation. As an inspiring approach for solving a variant of MTTDVRP, Pan et al. \cite{pan2021multi} introduced three functions: a time-dependent travel time function, a time-dependent ready time function, and a time-dependent duration function, for feasibility evaluation of the routing problem. This study introduced an innovative and efficient method to derive them through an iterative process. The authors proposed a hybrid algorithm that integrates the exploration capability of Variable Neighborhood Descent (VND) within a neighborhood with Adaptive Large Neighborhood Search (ALNS) to perturb solutions when VND becomes stuck in local optima. Zhao et al. \cite{zhao2024hybrid}, alternatively proposed another pioneering hybrid approach, combining a dynamic programming-based algorithm to partition a sequence of customer stops—a "giant tour"—into feasible vehicle trips. Additionally, they developed a hybrid genetic algorithm (GA), which integrates GA with a local search procedure to efficiently construct these giant tours.

In one of the pioneering studies using a DRL-based method to solve VRPs, Nazari et al. \cite{nazari2018reinforcement} proposed a DRL algorithm with a policy network consisting of an RNN-based decoder only network to address the basic VRP, training the policy model using an aggregated Euclidean distance metric as the feedback signal. In a subsequent study, Kool et al. \cite{kool2018attention} proposed a Transformer-based policy network trained with a self-critic reinforcement learning (RL) algorithm, achieving performance that surpassed established benchmarks, including Google OR-Tools.

Duan et al. \cite{duan2020efficiently} highlighted the ineffectiveness of previous studies, such as those by Nazari et al. and Kool et al., for real-world logistics operations where the objective is to minimize total road network travel distance or time, rather than total Euclidean-based distance or time. Alternatively, the proposed methods introduced a DRL-based algorithm called GCN-NPEC, featuring a GCN-based encoder and two decoders: one gated recurrent unit (GRU) and one multi-layer perceptron (MLP) as its policy network. GCN-NPEC successfully outperformed the competitive baseline, Google OR-Tools, in solving the VRP with the objective of minimizing the total road network travel distance/time. However, due to the use of GRU units, the model lacks parallelism and has inferior computational efficiency compared to the Transformer-style architecture employed by Kool et al. Alternatively, Zhang et al. \cite{zhang2022edge} proposed an edge selection-based approach called GAT-Edge, which utilizes the real traveling distance between nodes in its objective function rather than Euclidean distances by leveraging a Transformer-based architecture. GAT-Edge outperformed the AM, GCN-NPEC, and OR-Tools. Node selection-based methods, such as AM and GCN-NPEC, are of linear order with respect to the routing problem size. However, GAT-Edge at each decoding step outputs a probability distribution for the edges, which is of quadratic order of problem size, leading to significantly higher computational time, particularly for large-scale problems. Lei et al. \cite{lei2022solve} introduced an alternative method, named residual E-GAT, that utilizes edge information along with node-based information for solving the basic VRP. As E-GAT is a node-selection-based method, it avoids the aforementioned drawbacks of GAT-Edge. Nonetheless, GAT-Edge faces limitations due to its use of concatenation and inability to update edge information at each attention layer, resulting in reduced representational power. Although these three methods leverage edge information in their objective function rather than relying on the unrealistic estimation of traveling distance via Euclidean distance, they fail to account for the dynamic nature of traffic—and, consequently, the dynamic nature of travel time or optimal travel distance—which is a crucial factor in real-world road networks.

\section{Preliminaries} 
\label{section:preliminaries}

In this section, we present the preliminaries for MTTDVRP with a maximum working hours constraints, describe the problem, and introduce the MDP-based formulation. For ease of reference, a list of notations used throughout this paper is provided in Appendix \ref{appendixa}.

\begin{definition} \label{def:graph}
\textbf{(Customer-Depot Graph).} A customer-depot graph $G$ is a directed and complete graph defined as $G=(V, E)$, where $V=\{v_0, \dots, v_{n}\}$ and $E=\{e_{i,j} \mid 0 \leq i, j \leq n, i \neq j\}$ represent the sets of nodes and edges of the graph, respectively, and $n$ denotes the number of customers. The set $V_c = V \setminus \{v_0\}$ comprises the $n$ customer nodes, while the node $v_0$ denotes the depot.
\end{definition}

In MTTDVRP with the maximum working hours constraints, the traveling time between locations varies with the vehicles' departure time, so a time-dependent travel cost function is defined to represent these varying travel times.

\begin{definition} \label{def:cost}
\textbf{(Time-Dependent Travel Cost Function).} Given the set of nodes $V$ of the customer-depot graph $G$ and a time-interval set $TI=\{1,2,\dots,|TI|\}$, a time-dependent travel cost function $c: \mathbb{R}^{|V|\times |V|\times|TI|} \to \mathbb{R}$ is defined, where $|TI|$ is the length of the time-interval set. $\forall v_{i}, v_{j} \in V$, the value of $c(v_{i}, v_{j}, p)$ represents the travel cost, i.e., the travel time, associated with traveling from $v_i$ to $v_j$ during the time interval $p \in TI$.
\end{definition}

\begin{definition} \label{def:nodefv}
\textbf{(Node-based Feature Vector).} 
Given the customer-depot graph $G$'s node set $V$, a node-based feature vector $\mathcal{X} = \{(x_{v_0}, y_{v_0}, d_{v_0}), \dots, (x_{v_n}, y_{v_n}, d_{v_n})\}$ is defined. For each element $\chi_{v_i} \in \mathcal{X}$, corresponding to a node $v_i$, $(x_{v_i}, y_{v_i})$ are the 2-dimensional location coordinates of the node, and $d_{v_i}$ indicates the demand at that node (with $d_{v_0}$ set to 0).
\end{definition}

\begin{definition} \label{def:edgefv}
\textbf{(Edge-based Feature Tensor).} Given the edge set $E$ of the customer-depot graph $G$ and the time-dependent travel cost function $c$, an edge-based feature tensor $\mathcal{E} = \{c(v_{i},v_{j},p) | v_{i}, v_{j} \in V, p \in TI\}$ is defined to represent the travel time from $v_{i}$ to $v_{j}$ during the time-interval $p$. Each element ${\varepsilon_{ij,p}}$ of this tensor is equal to $c(v_{i}, v_{j}, p)$.
\end{definition}

In the MTTDVRP with maximum working hours, the objective is to minimize total travel time using a fleet of vehicles $F = \{1, \dots, f\}$, while serving customer $V_c$'s demands. Each vehicle has a maximum capacity of $Q$, starts working at time 0, and must return to the depot before $T_{max}$, which denotes the end of the workday. The day is divided into $|TI|$ time intervals, with travel time remaining constant within each interval. The vehicles, denoted by $(k,r)$ for the $k$-th vehicle's $r$-th trip, can undertake multiple trips per day, where $R_k$ represents the set of possible trips for vehicle $k$.

Solving the MTTDVRP with the maximum working hours constraints in a constructive manner can be deemed as a sequential decision-making problem. This process is modeled as a Markov Decision Process (MDP), represented by a 4-tuple $\{S, A, \mathcal{T}, r\}$, where the elements denote the state space, action space, a transition function, and a reward function, respectively, each defined as follows:
\begin{itemize}
  \item \textbf{State:} 
  The state of the system at step $t$ is represented by $s^t=(s_{F}^t,s_{R}^t) \in S$. The first element, $s_{F}^t$, corresponds to the fleet of vehicles' state. This element is expressed by $s_{F}^t=\left\{s^{t}_{1}, \dots, s^{t}_{f}\right\}=\left\{\left[rc_1^t, \mathcal{V}_1^t, \tau_1^t, p_1^t, rt_1^t \right], \dots, \left[rc_f^t, \mathcal{V}_f^t, \tau_f^t, p_f^t, rt_f^t\right]\right\}$.
  Here, $rc_i^t$ indicates the remaining capacity and $\mathcal{V}_i^t$ denotes the current location of vehicle $i$. $\tau_i^t$ is the vehicle's remaining working hours. $p_i^t$ refers to the current time-interval for the vehicle, and $rt_i^t$ is the time left in this interval. The second part, $s^t_{R}$, shows the routing state, represented by $s_{R}^t=\left\{V^t_{visited}, V^t_{to \textunderscore visit}\right\}$. $V^t_{visited}$ lists all customer nodes visited so far, and $V^t_{to \textunderscore visit}$ includes all nodes yet to be served. 
  \item \textbf{Action:} The action at step $t$ is expressed by a 2-tuple $a^t=(i^t, {v_j}^t) \in A$, indicating that node $v_j$ is visited by vehicle $i$ at step $t$. $A$ denotes the action space.
  \item \textbf{Transition:} The state transition function, denoted as $s_{t+1} = \mathcal{T}(a_t, s_t)$, updates the current system state $s_t$ to the next state $s_{t+1}$ based on the last taken action $a_t$. Given a vehicle $k$ and a node $v_j$ selected by the policy at step $t$, the state transition function updates every five elements of each vehicle $i$'s state according to the following rules:
\begin{eqnarray}  \label{form:trans:r1}
    rc_i^{t+1}= 
\begin{cases}
    Q,& \text{if } i = k =0\\
    rc_i^t-d_{v_j},& \text{if } i = k \neq 0\\
    rc_i^{t},              & \text{otherwise},
\end{cases}
\end{eqnarray}
\begin{eqnarray} \label{form:trans:r2}
    \mathcal{V}_i^{t+1}= 
\begin{cases}
    v_j,& \text{if } i = k\\
    \mathcal{V}_i^{t},              & \text{otherwise},
\end{cases}
\end{eqnarray}
\begin{eqnarray} \label{form:trans:r3}
    \tau_i^{t+1}= 
\begin{cases}
    \tau_i^{t}-c({V}_i^{t},v_j,p_i^{t}),& \text{if } i = k\\
    \tau_i^{t},              & \text{otherwise},
\end{cases}
\end{eqnarray}
\begin{eqnarray} \label{form:trans:r4}
    p_i^{t+1}= 
\begin{cases}
    p_i^{t}+ \left\lceil \frac{T_{max} - \tau_i^{t+1}}{T_{max}} \times |TI|\right\rceil,& \text{if } i = k \text{ and } T_{max} - \tau_i^{t+1} \geq tw^{p_i^{t}+1} \\
    p_i^{t},              & \text{otherwise},
\end{cases}
\end{eqnarray}
\begin{eqnarray} \label{form:trans:r5}
    rt_i^{t+1}= 
\begin{cases}
    tw^{p_i^{t}+1}-tw^{p_i^{t}}+rt_i^{t}-c({V}_i^{t},v_j,p_i^{t}),& \text{if } i = k \text{ and } T_{max} - \tau_i^{t+1} \geq tw^{p_i^{t}+1} \\
    rt_i^{t}-c({V}_i^{t},v_j,p_i^{t}),              & \text{if } i = k \text{ and } T_{max} - \tau_i^{t+1} < tw^{p_i^{t}+1} \\
    rt_i^{t},              & \text{otherwise},
\end{cases}
\end{eqnarray}
Rule \ref{form:trans:r1} reduces the selected vehicle's remaining capacity by the customer node's demand, or resets it, for the depot node.
Rule \ref{form:trans:r2} updates the vehicle's current location to the node selected at step $t$. The rule \ref{form:trans:r3} adjusts the remaining working hours of the selected vehicle $i$'. Rule \ref{form:trans:r4} determines the vehicle's current time interval, bounded by the period $\left[tw^{p_i^{t}}, tw^{p_i^{t}+1}\right]$, based on its working hours. Finally, Rule \ref{form:trans:r5} modifies the vehicle's remaining time in the current interval.
Similarly, if $j \neq 0$ the two elements of the routing state would be updated as follows:
\begin{eqnarray}
V^{t+1}_{visited} = V^{t}_{visited} \cup \{v_j\}, \quad \quad \quad \quad \quad V^{t+1}_{to \textunderscore visit} = V^t_{to \textunderscore visit} \setminus \{v_j\}.
\end{eqnarray}
  \item \textbf{Rewards:} For each step $t$, a single-step reward is defined as follows:
\begin{eqnarray}  
r(s^{t},a^{t}) = -c(\mathcal{V}_i^{t-1}, \mathcal{V}_i^{t}, p_i^{t-1}).
\end{eqnarray}
Therefore, given a solution $\Pi$, the aggregated reward function $R$ can be expressed as
\begin{eqnarray} \label{form:reward} 
R(\Pi)= \sum_{t = 1}^{T} r(s^{t},a^{t}),
\end{eqnarray}
where $T$ denotes the number of decision steps in a complete episode.
\end{itemize}

\section{Methodology} \label{section:method}

In this section, we introduce our Simultaneous Encoder and Dual Decoder Attention Model (SED2AM) for solving the MTTDVRP with a maximum working hours constraints. First, we provide a general framework description, followed by a detailed explanation of the Transformer-style policy model for the DRL method and conclude with the training procedure of the policy model.

\subsection{Framework Overview} \label{policy}

\begin{figure}[h!]
    \includegraphics[width=5.5in]{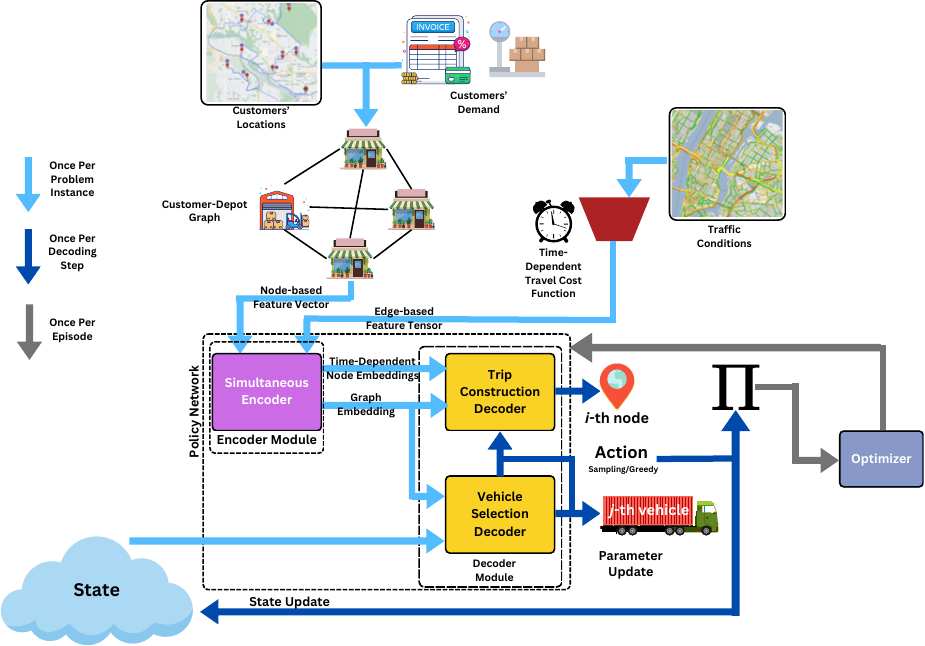}
  
  \caption{Overall architecture of SED2AM}
  \label{diagram:framework}
\end{figure}

Figure \ref{diagram:framework} illustrates the overall architecture of the SED2AM model, a framework designed for the route construction task for the MTTDVRP with maximum working hours constraints. The route construction task is formulated as an episodic reinforcement learning problem and represented by a Transformer-based model parameterized by trainable parameters $\theta$. For each MTTDVRP instance to be solved, the preprocessing procedure starts with constructing the customer-depot graph $G$, as defined in Definition \ref{def:graph}, to construct the node-based feature vector $\mathcal{X}$ (defined in Definition \ref{def:nodefv}), as depicted in Figure \ref{diagram:framework}. 
Concurrently, the edge-based feature tensor $\mathcal{E}$ (also defined in Definition \ref{def:edgefv}) is constructed using the time-dependent travel cost function $c$. Subsequently, the node-based feature vector and the edge-based feature tensor are passed to the policy model to solve the MTTDVRP instance in a constructive manner. As perceivable through Figure \ref{diagram:framework}, SED2AM's policy model consists of an encoder and two decoders for vehicle selection, and route construction.

The process of solving a problem instance, i.e., route construction, by the policy model begins with an encoding step where the constructed node-based feature vector $\mathcal{X}$ and the edge-based feature tensor $\mathcal{E}$ are initially passed to the policy model. This encoding process is followed by an episode comprised of $T$ decoding steps. Each decoding step starts at the vehicle selection decoder, which selects a vehicle based on the current state of the fleet. The route construction decoder then chooses the node to be visited by the selected vehicle. This decoding process concludes after $T$ steps when all customers have been served. In each episode, given an initial state $s^0$ and a terminal state $s^T$, the conditional probability of the constructed solution can be expressed, under the Markov property, as:

\begin{eqnarray} \label{eq:policy}
p_\theta(\Pi|s_0)=\prod_{t=0}^{T-1}\pi_\theta(a^t|s^t)\mathcal{T}(s^{t+1}|a^t,s^t).
\end{eqnarray}
where $\mathcal{T}(s^{t+1}|a^t,s^t)$ is the deterministic state transition function described in Section \ref{section:preliminaries}, and $\Pi$  denotes the solution to the problem instance.

At the end of each episode, the total cost, which is the aggregated time-dependent travel time for the constructed trips, is computed for the solution $\Pi$. This total cost is then used as feedback for a policy-gradient-based reinforcement learning algorithm (explained in Section \ref{training}). The optimizer within the reinforcement learning algorithm subsequently updates the policy model parameters, $\theta$, based on this feedback.

\subsection{Policy Model Architecture} \label{sec:architecture}

SED2AM uses a Transformer-style policy model for route construction to solve MTTDVRP with a maximum working hours constraints. As perceivable in Figure \ref{diagram:framework}, the policy network includes three main components: 1) simultaneous encoder, 2) vehicle selection decoder, and 3) trip construction decoder. The following sections detail each component.

\subsubsection{Simultaneous Encoder} For a given routing problem instance, represented by the customer-depot graph $G$, the encoder module of SED2AM simultaneously co-embeds features attributed to both nodes, i.e. 2-D coordinates and customers' demand, and edges, i.e. time-dependent traveling time, of the customer-depot graph. The encoder outputs a high-dimensional latent representation of the problem instance. These problem instance embeddings, as elements of the Transformer-style policy network's state representation, play an important role in effectively representing the routing problem instance. To effectively reflect the variability in travel time within the state representation, the simultaneous encoder independently embeds each time-interval. This design choice stems from the understanding that, at each decoding step, the selection of the next destination for any given vehicle should be solely contingent upon the traffic conditions experienced during that specific time-interval.

\begin{figure}[h!]
    \includegraphics[width=4.7in]{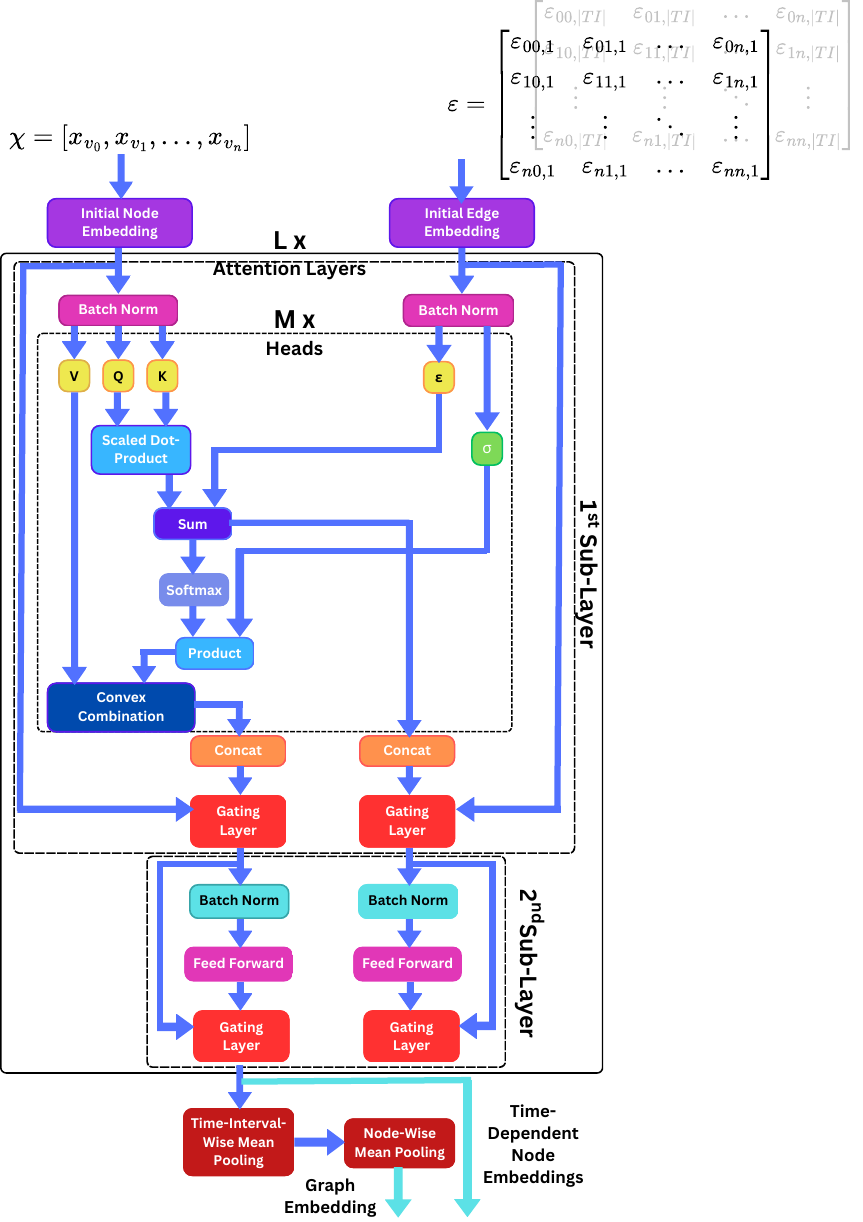}
  
  \caption{Illustration of the simultaneous encoder architecture: the encoder computes time-dependent node embeddings by simultaneously co-embedding the node features and edge features through $L$ layers of attention.}
  \label{diagram:encoder}
\end{figure}

The encoding process, as illustrated in Figure \ref{diagram:encoder}, begins with computing the initial embeddings of the node features and edge features. For a given node feature vector $\mathcal{X}$ (defined in Definition \ref{def:nodefv}), the initial node embedding for each node $v_i$ during time-interval $p$ is calculated using a linear projection with adjustable parameters $W^\chi$ and $b^\chi$, as in $h_{v_i,p}^{0}=W^{\chi}{\chi_{v_i}}+b^\chi$. Similarly, given the edge feature vector $\mathcal{E}$, the initial edge embedding for each edge $e_{ij}$ during the time-interval $p$ is determined using a linear projection with a different set of adjustable parameters, represented by $h_{e_{ij,p}}^{0}=W^{\mathcal{E}}\varepsilon_{ij,p}+b^\mathcal{E}$. This initial embedding step, which uses linear projections for both node and edge features, is critical for mapping these features into a high-dimensional vector space. This alignment ensures compatibility in dimensionality, enabling effective self-attention calculations within the Transformer architecture. Furthermore, this embedding design enables the self-attention mechanism to identify and leverage semantic relationships within the graph, enhancing the model’s ability to capture dependencies between nodes based on both node and edge features, thereby facilitating effective routing for the given problem instance.

Following that, the initial node and edge embeddings are processed through $L$ attention layers, each consisting of two sub-layers. The first sub-layer simultaneously co-embeds the node-related and edge-related inputs using a multi-head attention layer. In the following sub-layer, node-related and edge-related outputs of the previous sub-layer are separately passed to two different fully-connected feedforward networks. 

In each attention layer $l$, we first normalize the node-related and edge-related inputs using batch normalization, to compute normalized embeddings $h^{l}_{BN,v_i,p}$ and $h^{l}_{BN,e_{ij},p}$, respectively. Batch normalization aids in stabilizing the training process and mitigating the risk of exploding or vanishing gradients \cite{pmlr-v37-ioffe15}.

Then, the $key$, $query$, and $value$ combination for the $m$-th attention head for each time-dependent node-related input embedding $h^l_{v_i,p}$ are respectively computed as $q^{l,m}_{i,p} = W_{Q,p}^{l,m}h^{l}_{BN,v_i,p}, k^{l,m}_{i,p} = W_{K,p}^{l,m}h^{l}_{BN,v_i,p}, \nu^{l,m}_{i,p} = W_{V,p}^{l,m}h^{l}_{BN,v_i,p}$, where $W_{Q,p}^{l,m}$,$W_{K,p}^{l,m}$, and $W_{V,p}^{l,m}$ are trainable parameters. Similarly, for each attention layer $l$ and attention head $m$, using trainable parameters $W_E^{l,m}$, a linear projection of the edge input for the time-interval $p$ is calculated as $\epsilon^{l,m}_{ij,p} = W_{E,p}^{l,m}h^{l}_{BN,e_{ij},p}$. The individual projections for queries, keys, values, and edge features enable the model to learn distinct representations for each component. Multiple attention heads allow the model to capture diverse dependencies within the data by enabling each head to focus on different aspects of relationships in the graph. In SED2AM, the encoder applies the attention mechanism independently for each time interval $p$. This design introduces a temporal locality inductive bias, meaning that each decision-making step depends solely on the traffic conditions during the specific interval in which the vehicle is traveling to its next destination. The rationale for this design is that each routing decision is influenced only by immediate, relevant traffic conditions during travel, making it independent of traffic conditions at other times.

To effectively quantify the influence of each node’s query and key values within the customer-depot graph, and subsequently to modulate the value received by each node $v_i$, our model employs the $compatibility$ score $u^{l,m}_{ij,p}$. This score is computed as follows:

\begin{eqnarray} \label{eq:compat}
u^{l,m}_{ij,p} = \dfrac{{q^{l,m}_{i,p}}^\intercal k^{l,m}_{j,p}}{\sqrt{d_k}}+\epsilon^{l,m}_{ji,p}.
\end{eqnarray}

where $d_k$, denoting the $key$ dimension per attention head, scales the dot product between $query$ and $key$ vectors to normalize the output and stabilize the attention weights. This scaling mitigates excessively large  $compatibility$ score values as dimensionality increases, thereby preventing gradient instability and aiding optimization \cite{vaswani2017attention}.

The incorporation of edge information through the addition of edge embedding $\epsilon^{l,m}_{ji,p}$ modulates the base attention scores in accordance with the dynamics of traffic conditions, allowing the model to make informed decisions based on time-dependent traveling time between the nodes as a biased adjustment in addition to the inherent attributes of the nodes. The summation operation offers a straightforward yet effective way to incorporate edge embeddings into  $compatibility$  scores. By linearly combining node and edge information, summation preserves the original dimensionality, avoiding the need for additional transformations or increased parameter complexity \cite{gete2023evaluation}. This simplicity promotes stable gradient behavior, facilitating smooth optimization in high-dimensional attention layers. Furthermore, summation prevents abrupt changes in $compatibility$ scores, allowing the model to adapt smoothly to varying edge information without introducing instability \cite{kwon2020motionsqueeze}.

Then, using the $compatibility$ $u^{l,m}_{ij,p}$ and the edge embedding $\epsilon^{l,m}_{ji,p}$, the $attention$ $weight$ is computed using the following equation:
\begin{eqnarray} \label{eq:att}
a^{l,m}_{ij,p} = \dfrac{e^{u^{l,m}_{ij,p}}}{\sum_k e^{u^{l,m}_{ik,p}}} \times \sigma\left(\epsilon^{l,m}_{ji,p}\right).
\end{eqnarray}
where $\sigma$ denotes the Sigmoid function. The softmax function in Equation \ref{eq:att} normalizes the compatibility scores across all nodes, converting them into a probability distribution that emphasizes the most compatible nodes. 

By bounding the edge weights, the Sigmoid function moderates the impact of edge-related values on the attention weights, ensuring they remain within the Sigmoid function's $[0,1]$ range. Equation \ref{eq:att} integrates the attention probabilities (i.e., softmaxed compatibility scores) and edge features, adjusting the significance of nodes relative to each other based on the dynamic, time-dependent traffic conditions.

Then, a convex combination of values $\nu^{l,m}_{j,p}$ passed to each node $v_i$ is computed as follows:
\begin{eqnarray} \label{eq}
{h^\prime}_{v_i,p}^{l,m} = \sum_ja^{l,m}_{ij,p}\nu^{l,m}_{j,p}.
\end{eqnarray}
For each time-interval $p$ and node $v_i$, the final output of the multi-head attention (MHA), utilizing a single attention head with parameter matrices $W_{O,p}^{l,m}$, is given by $MHA({h}_{v_i,p}^{l}) = \sum_{m=1}^{M}W_{O,p}^{l,m}{h^\prime}_{v_i,p}^{l,m}$. This step synthesizes information across multiple representational subspaces, thereby enhancing the decision-making process by enabling the model to capture the diverse underlying relationships within the customer-depot graph under varying traffic conditions.

We use the gating mechanism proposed in \cite{parisotto2020stabilizing} to compute the final node-related, $g^l_{v_i,p}$, and edge-related, $g^l_{e_{ij},p}$, results of co-embedding by the first sub-layer of the attention layer as follows:
\begin{eqnarray} \label{eq}
g^l_{v_i,p}=MHA
({h}_{v_i,p}^{l}) \cdot z^l_{v_{i,p}} + {h}_{v_{i,p}}^{l},
\end{eqnarray}
\begin{eqnarray} \label{eq}
g^l_{e_{ij},p}=FF\left(u^\prime{^l_{ij,p}} || \epsilon^{l,m}_{{ij},p}\right) \cdot z^l_{e_{ij},p} + h^{l}_{e_{{ij},p}},
\end{eqnarray}
where $FF$ and $||$ respectively represent a feed-forward layer with linear activation function and concatenation operation, and  
\begin{eqnarray} \label{eq}
z^l_{v_i,p} = \sigma\left(W^l_{G_v,p}h^{l}_{v_i,p}+b^l_{G_v,p}\right), \quad \quad z^l_{e_{ij},p} = \sigma \left(W^l_{G_{e,p}}h^{l}_{e_{ji},p}+b^l_{G_e,p}\right),
\end{eqnarray}
where $W^l_{G_{v,p}}$ and $W^l_{G_{e,p}}$ are learnable weights. $b_{G_{v,p}}$ and $b_{G_{e,p}}$ are learnable biases. Parisotto et al. \cite{parisotto2020stabilizing} demonstrated that this gating mechanism stabilizes learning and enhances performance when training Transformer models with reinforcement learning. This is the rationale for its use in the simultaneous encoder.

Last, the node-related, i.e $h^{l+1}_{v_i,p}$, and edge-related, i.e $h^{l+1}_{e_{ij},p}$, embeddings output on the $l$-th layer are computed after passing through the second sub-layer through the following equations:
\begin{eqnarray} \label{eq:node_embed}
h^{l+1}_{v_i,p}=FF\left(BN\left(g^l_{v_i,p}\right)\right)\cdot z^\prime{^l_{v_i,p}}+g^l_{v_i,p},
\end{eqnarray}
\begin{eqnarray} \label{eq:node_embd_encoder}
h^{l+1}_{e_{ij},p}=FF\left(BN\left(g^l_{e_{ij},p}\right)\right)\cdot z^\prime{^l_{e_{ij},p}}+g^l_{e_{ij},p},
\end{eqnarray}
where $FF$ denote feed-forward sub-layers with ReLU activation function, and
\begin{eqnarray} \label{eq}
{z^\prime}^l_{v_i,p} = \sigma\left(W^l_{H_{v,p}}g^l_{v_i,p}+b^l_{H_{v,p}}\right), \quad \quad {z^\prime}^l_{e_{ij,p}} = \sigma\left(W^l_{H_{e,p}}g^l_{e_{ij},p}+b^l_{H_{e,p}}\right),
\end{eqnarray}
where $W^l_{H_{v,p}}$ and $W^l_{H_{e,p}}$ are learnable weights, and $b^l_{H_{v,p}}$ and $b^l_{H_{e,p}}$ are learnable biases. The feed-forward sub-layers perform a non-linear transformation, augmenting the encoder's representation power and enabling it to capture more complex patterns in the data.

For each problem instance, the encoder module produces time-dependent node-embeddings $h^L_{v_i,p}$ for each node $v_i$ and time interval $p$ after $L$ layers of attention. We simplify notation by dropping the superscript $L$, using $h_{v_i,p}$ instead. The graph embedding for each instance is the average over all time-intervals and nodes, calculated as $\frac{1}{|TI| \times n}\sum_{p\in TI} \sum_{i=0}^{n} h_{v_i,p}$, where $|TI|$ is the number of time-intervals.

\subsubsection{Vehicle Selection Decoder} \label{decoder1}
The vehicle selection decoder, at each decoding step, selects a vehicle that has not reached its maximum working hours to either expand its current trip or start a new one. This decoder outputs a probability distribution over all vehicles to be chosen based on the current state of the fleet. Such an expansion of the action space facilitates a more effective and rational policy search, leading to improved control policies and better routing performance.

\begin{figure}[h!]
    \includegraphics[width=3.5in]{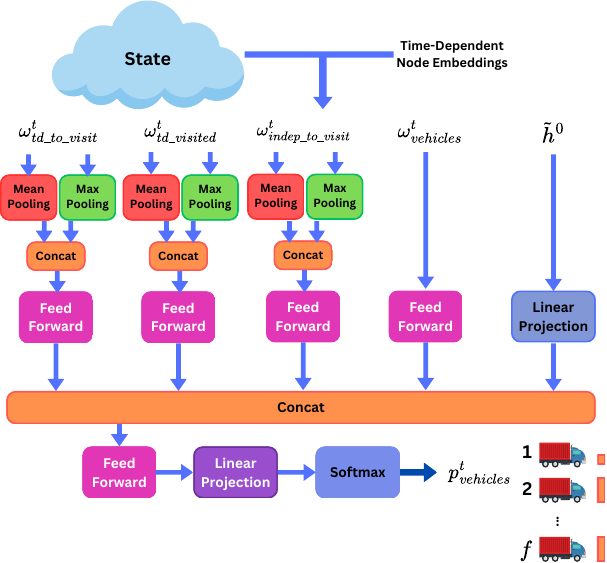}
  \caption{At each decoding step $t$, the vehicle selection decoder generates a probability distribution over the vehicles, selecting one to expand its route.}
  \label{diagram:vehicle_decoder}
\end{figure}

Figure \ref{diagram:vehicle_decoder} illustrates the vehicle selection decoder model within the SED2AM framework, which calculates a probability distribution for selecting a vehicle from the fleet. To compute these probabilities, the decoder model receives five input elements, each described as follows:

\begin{enumerate}
    \item Vehicle status embedding: This embedding provides a critical representation of the current state of the vehicles, effectively capturing the state transition dynamics essential for informed routing \cite{xu2021reinforcement}. At each step $t$, the vehicle status feature vector is derived from the system state representation, as detailed in Section \ref{section:preliminaries}, through the following equation:
    
\begin{eqnarray} \label{eq:tsc_embd}
\omega^t_{vehicles}=\left\{s_1^{t}, \dots, s_f^{t}\right\}=\left\{\left[rc_1^t, \mathcal{V}_1^t, \tau_1^t, p_1^t, rt_1^t\right], \dots, \left[rc_f^t, \mathcal{V}_f^t, \tau_f^t, p_f^t, rt_f^t\right]\right\}
\end{eqnarray}
where $f$ denotes to the number of vehicles in the fleet. $rc_i^t$ ensures the model makes informed decisions while adhering to the vehicle’s maximum capacity constraints. Similarly, $\tau_i^t$ allows the model to select the next vehicle effectively based on the maximum working hours constraints. $p_i^t$ and $rt_i^t$ enable the model to capture and adapt to the dynamics of traffic conditions. At each step $t$, the vehicle status embedding is computed as follows:

\begin{eqnarray} \label{eq}
e^t_{vehicles}=FF\left(\omega^t_{vehicles}\right),
\end{eqnarray}

$FF$ denotes a two-layer feed-forward neural network in which the first hidden layer uses a linear activation function, and the second layer uses a ReLU activation function. This neural network embeds and reduces the dimensionality of the input while introducing non-linearity to enhance expressiveness.

\item Visited customers' time-dependent embedding: This embedding plays a crucial role in enabling the decoder model to learn from its past actions by capturing information about the travel to customer nodes was done, segmented according to the specific time intervals during which service was provided. This information is crucial for conducting effective policy searches, as it incorporates the dynamics of action selection, thereby enhancing the decision-making process \cite{xu2021reinforcement}.
At each step $t$, we define a visited customer set $V^t_{visited} \subset V_c$ as the set of customers that have been served up this point. We additionally define the visited customers' time-dependent feature vector $\omega^t_{td \textunderscore visited} \in \mathbb{R}^{dim \times |V^t_{visited}|}$ (where $dim$ corresponds to the size of the hidden dimension and $|V^t_{visited}|$ denotes the number of customer nodes that have been served so far), as follows:
\begin{eqnarray} \label{eq:td_emb}
\omega^t_{td \textunderscore visited  }=\left[\Ddot{h}^i_{td}|i \in V^t_{visited}\right],
\end{eqnarray}
It is noteworthy that $|V^t_{visited}| \leq t$, as the depot can be visited multiple times by the vehicles in a working day, and $\Ddot{h}^i_{td}$ represents the node embedding for the $i$-th customer at the corresponding time-interval that it has been visited (defined in Equation \ref{eq:node_embed}). Visited customers' time-dependent embedding $e^t_{td \textunderscore visited}$ at step $t$ is then computed by concatenating the results of max pooling and mean pooling operations on the vector $\omega^t_{td \textunderscore visited}$. This procedure is outlined as follows: 
\begin{eqnarray} \label{eq}
e^t_{td \textunderscore visited}=FF\left(\max \left(\omega^t_{visited \textunderscore td}  \right) || \left(\frac{1}{|\omega^t_{visited}|} \sum_{i \in V^t_{visited}} \Ddot{h}^i_{td}\right)\right),
\end{eqnarray}
where $FF$ is a two-layer feed-forward neural network. The first hidden layer uses a linear activation function, while the second employs a ReLU activation function to introduce nonlinearity. The hidden dimension of the feed-forward network is set to $2 \times \text{dim}$.

\item To-be visited customers' time-dependent embedding: This embedding equips the policy network with critical information to evaluate the available options for the vehicle selected at decoding step $t$. For each step $t$, given the time-interval $p_j^t, j\in{1,\dots,f}$ that each vehicle $j$ is experiencing, the time-dependent node embeddings set ${h_{i,p} | i \in V, p \in TI}$, customers' set $V_c$, and visited customers set $V^t_{visited}$, the to-be visited customers’ time-dependent feature vector is defined as

\begin{eqnarray} \label{eq}
    \omega^{t,p}_{td \textunderscore to \textunderscore visit}=\left\{h_{i,p}|i \in V_c \setminus V^t_{visited}\right\},
\end{eqnarray}
Then, the to-be visited customers' time-dependent embedding $\omega^t_{td \textunderscore to \textunderscore visit}$ is computed as follows:
\begin{eqnarray} \label{eq:tsc_embd}
e^t_{td \textunderscore to \textunderscore visit}=FF\left(\max \left(\omega^{t,p}_{td \textunderscore to \textunderscore visit}  \right) || \left(\frac{1}{|\omega^{t,p}_{td \textunderscore to \textunderscore visit  }|} \sum_{i\in V_c \setminus V^t_{visited}} h_{i,p}\right)\right),
\end{eqnarray}
where  $FF$ denotes a two-layer feed-forward neural network with a hidden dimension of size $2 \times dim$ and linear activation function for the first layer and ReLU for the second one. This embedding is utilized to empower the policy model to perform the vehicle selection based on the traffic conditions that each of the vehicles experiences at the current decoding step.

\item To-be visited customers’ time-independent embedding: This embedding provides a representation of the remaining unserved customers, independent of the current state of the vehicles. In conjunction with the visited customers’ time-dependent embedding, it enables the model to adapt to changes across decoding steps effectively. 
For each step $t$, given the visited customers set $V^t_{visited}$, the to-be-visited customers’ time-independent feature vector is defined as

\begin{eqnarray} \label{eq}
    \omega^t_{indep \textunderscore  to \textunderscore visit}=\left\{{\Tilde{h}}^i|i \in V_c \setminus V^t_{visited}\right\},
\end{eqnarray}
To-be visited customers’ time-independent embedding $e^t_{to \textunderscore visit \textunderscore indep}$ is then computed through the following equation:
\begin{eqnarray} \label{eq:tsc_embd}
e^t_{indep \textunderscore  to \textunderscore visit }=FF\left(\max \left(\omega^t_{indep \textunderscore  to \textunderscore visit}  \right) || \left(\frac{1}{|\omega^t_{indep \textunderscore  to \textunderscore visit}|} \sum_{i\in V_c \setminus V^t_{visited}} {\Tilde{h}}^i\right)\right),
\end{eqnarray}
where $FF$ represents a two-layer fully-connected network with a linear activation function in the first layer and ReLU in the second. This embedding empowers the model to select the next vehicle by considering the remaining customers to be served and factoring in potential traffic conditions when they are served in the following decoding steps.

    \item Depot's embedding: Given the embedding $\Tilde{h}^0$, the depot's embedding is computed as follows:
\begin{eqnarray} \label{eq:tsc_embd}
e_{depot}=\left(W_{depot} \Tilde{h}_0 + b_{depot}\right),
\end{eqnarray}
where $W_{depot}$ and $b_{depot}$ are the parameters for a linear projection. The depot's embedding, in combination with the embeddings for to-be-visited customers (time-independent and time-dependent) and visited customers (time-dependent), provides the model with critical structural information about the customer-depot graph. This combined embedding framework offers the model a comprehensive perspective on customer locations relative to the depot, incorporating both static features (coordinates) and dynamic conditions (time-dependent travel times). By integrating these elements, the model gains a well-rounded understanding of the problem instance, allowing for more informed decision-making.

\end{enumerate}

At each decoding step $t$, given the visited customers' time-dependent embedding $e^t_{visited \textunderscore td}$, the visited customers' time-independent embedding $e^t_{visited \textunderscore indep}$, the to-be visited customers' embedding $e^t_{to \textunderscore visit}$, the depot's embedding $e_{depot}$, and the vehicle status embedding $e^t_{vehicles}$, we compute the probability vector of vehicles $p^t_{vehicles}$ via a two-layer fully-connected network, with a linear activation function for the first layer and ReLU for the second layer, followed by a softmax function:
\begin{eqnarray} \label{eq}
h^t_{vehicles}=FF\left(e^t_{td \textunderscore visited} || e^t_{td \textunderscore to \textunderscore visit} || e^t_{indep \textunderscore to \textunderscore visit} || e_{depot} || e^t_{vehicles}\right),
\end{eqnarray}

\begin{eqnarray} \label{eq}
p^t_{vehicles}=softmax(h{^t_{vehicles}}).
\end{eqnarray}
$p^t_{vehicles} = [p^{t,1}_{vehicles}, \dots, p^{t,f}_{vehicles}]$ is a probability vector whose element $p^{t,i}_{vehicles}$ denotes the probability of selecting the vehicle $i$ to expand its routes at the $t$-th decoding step. The two-layer feedforward (FF) network enables the model to learn complex, non-linear relationships. This architecture allows the network to effectively capture both dynamic information (state dynamics influenced by the actions taken and traffic conditions) and static information (the structural characteristics of the problem instance) required to determine vehicle priorities at each decision step. The final softmax layer transforms the output of the FF network into a probability distribution over the vehicles.

\subsubsection{Trip Construction Decoder} \label{decoder2}

\begin{figure}[h!]
    \includegraphics[width=2.6in]{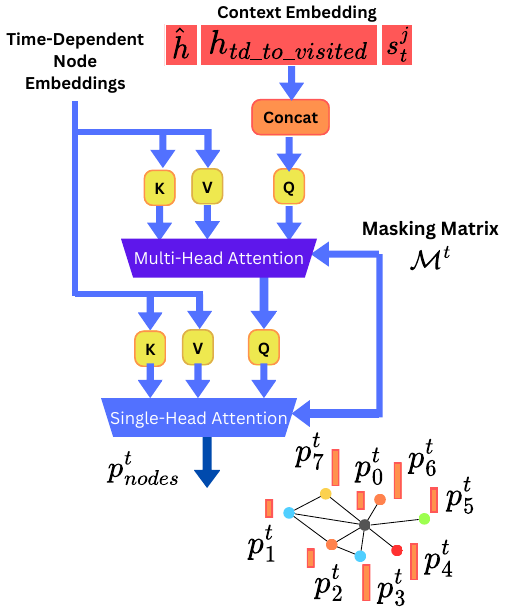}
  
  \caption{The architecture of the trip construction decoder. At each decoding step, this decoder outputs a probability distribution over unvisited nodes along with the depot based on the current state of the selected vehicle, graph embedding, and remaining nodes' time-dependent embeddings. Based on these probabilities, the node to be visited by the corresponding vehicle is then selected.}
  \label{diagram:route_decoder}
\end{figure}

At each decoding step $t$, for the vehicle $j$ chosen by the vehicle selection decoder at the current decoding step, the trip construction decoder outputs a probability distribution $p^t_{nodes}$ using the vehicle state vector $s_j^{t}$, (as defined in Section \ref{section:preliminaries}) and node embeddings from the encoder module (see Figure \ref{diagram:route_decoder}). This distribution determines the next node to be visited by vehicle $j$.

The trip construction decoding at each decoding step $t$ starts with the calculation of context node embedding $h^t_{(c)}$. Given the time-dependent to-be visited node embeddings $\hat{h}^t_{td \textunderscore to \textunderscore visit}$, the vehicle $j$ chosen by the vehicle selection decoder, the time-interval $p$ in which vehicle $j$ departs the last visited node, vehicle $j$'s state vector $s^{t}{j}$, and node embeddings $h{v_i,p}, v_i \in V$, the context node embedding $h^t_{(c)}$ is computed as follows:

\begin{eqnarray} \label{eq:context_embd}
 h^t_{(c)} = \left(\hat{h} || \hat{h}^t_{td \textunderscore to \textunderscore visit} || FF\left(s_j^{t}
 \right) \right)
\end{eqnarray}

where $||$ and $FF$ respectively represent the concatenation operation and feed-forward network with a linear activation function. The graph representation $\hat{h}$ in the contextual embedding enables the model to learn the structure of the customer-depot graph representing the problem instance and make informed routing decisions based on the given problem instance. The second element in the context embedding, i.e., time-dependent to-be visited node embeddings $\hat{h}^t_{td \textunderscore to \textunderscore visit}$, captures the dynamics of state transitions. It also reflects the current traffic conditions encountered by the chosen vehicle, reflecting the evolving traffic scenario. In the third element, while $\tau_j^t$ and $p_j^t$ enable the policy model to make informed decisions based on the dynamics of the traffic conditions, $rt_j^t$ provides it with information crucial for considering the maximum working hours constraints. $rc_j^t$ denotes the remaining capacity of the vehicle for consideration of the capacity constraint.

Given the context embedding $h^t_{(c)}$, the time-interval $p$, and time-dependent node embeddings $h_{p} = \left\{h_{v_{i},p}|v_i \in V_c\cup \{v_0\}\right\}$, using an MHA layer, the node embeddings' glimpse ${h^\prime}^t_{(c)}$ is computed as follows:

\begin{eqnarray} \label{eq:glimpse}
{h^\prime}^t_{(c)} = MHA(W^Q_{(c)}h^t_{(c)}, W^K_{(c)}h_{p}, W^V_{(c)}h_{p})
\end{eqnarray}

where $W^Q_{(c)}$, $W^K_{(c)}$, and $W^V_{(c)}$ are respectively the trainable parameters for $query$, $key$, and $value$. Utilizing the context embedding $h^t_{(c)}$, which represents the current state of the system, the glimpse mechanism, powered by multi-head attention, focuses on the relevant nodes pertinent to the system’s current conditions. Multi-head attention enables the model to analyze various relationships among the nodes, thereby facilitating informed decisions based on the state dynamics and traffic conditions.
Afterwards, given a node $v_i$ and time-interval $p$, the attention correlations between the time-dependent node embedding $h_{v_i,p}$ and the glimpse ${h^\prime}^t_{(c)}$ are computed using an attention function as follows:

\begin{eqnarray} \label{eq:kqv_comp_decoder2_tanh}
{u^\prime}^t_{(c),i} = 
\begin{cases}
    C.tanh\left(\dfrac{\alpha^t_{i}}{\sqrt{d_k}}\right),& \text{if } \mathcal{M}^{t}_{v_i} = 0 \\
    -\infty,& \text{otherwise.}\\
\end{cases}
\end{eqnarray} 

where

\begin{eqnarray} \label{eq:attnetionscore}
\alpha^t_{i} = \dfrac{{\left(W^\prime_Q{h^\prime}^t_{(c)}\right)}^\intercal \left(W^\prime_Kh_{v_i,p}\right)}{\sqrt{d_k}}
\end{eqnarray}

and $C$ is a constant that controls the entropy of ${u^\prime}^t_{(c),i}$, with its value set to 10. This layer, known as the compatibility layer, computes the attention scores for nodes based on the subset of nodes deemed relevant in the glimpse context \cite{bhunia2021more}. Given a customer node set $v_i$ and the selected vehicle $j$, the masking element $\mathcal{M}^{t}_{v_i}$ is defined as

\begin{eqnarray} \label{eq:mask}
    \mathcal{M}^{t}_{v_i} = 
\begin{cases}
    1,& \text{if } v_i \in V^t_{visited} \\
    1,& \text{if } \mathcal{V}_j^{t-1} = v_0 \text{ and }  \mathcal{V}_j^t = v_0\\
    1,& \text{if } \mathcal{V}_j^t \neq v_0 \text{ and } \tau_j^t < c(\mathcal{V}_j^{t}, v_i, p_j^{t}) + c(v_i, v_0, p_{Next})\\
    1,& \text{if } rc_j^t < d_{v_i}\\
    0,              & \text{otherwise,}
\end{cases}
\quad \quad \quad v_i \in V_c \cup \left\{v_0\right\},
\end{eqnarray}
where $p_{Next}$ corresponds to the time-interval that vehicle $j$ will reach node $v_i$ if the decoder selects $v_i$ as its next destination, and $\mathcal{M}^{t}_{v_i}=1$ indicates that $v_i$ shouldn't be visited at time-step $t$. The first rule in Equation \ref{eq:mask} ensures each customer node is visited once. The second rule makes sure that the depot is not visited twice at subsequent steps by the same vehicle. The third rule enforces the maximum working hours constraints. Lastly, the fourth rule disqualifies customers whose demands surpass the vehicle's available capacity.

Finally, at each decoding step $t$, a probability vector $p^{t}_{nodes}$ is calculated using a softmax function, where each element $p^{t,i}_{nodes}$, denoting the probability of the node $v_i$ being visited by vehicle $j$, is calculated as follows:
\begin{eqnarray} \label{eq:prob}
p^{t,i}_{nodes} = 
\begin{cases}
    \dfrac{{e^{u^\prime}}^t_{(c),i}}{\sum_k e^{{u^\prime}^t_{(c),i}}},& \text{if } \mathcal{M}^{t}_{v_l} = 0 \\
    0,& \text{otherwise.}\\
\end{cases}
\end{eqnarray}

\subsection{Computational Complexity Analysis}
In this section, the computational complexity of the Transformer-style policy network proposed by Kool et al. \cite{kool2018attention} is compared with that of SED2AM. Table \ref{tab:complexity} provides a comparison of the computational complexities of these two approaches, with notations defined in Table \ref{tab:complexity_notations}.

The simultaneous encoder employed by SED2AM introduces an additional multiplicative factor of \( |TI| \), representing the number of time intervals. However, \( |TI| \) is a fixed constant and does not scale with problem size. Therefore, in the context of asymptotic analysis, it can be treated as a constant factor and omitted from the \( O \)-notation for time complexity.

As detailed in Table 3, the inclusion of the vehicle selection decoder in SED2AM adds a computational complexity of \( O(n \cdot d \cdot f) \) for the decoding step. It is important to note that the fleet size \( f \) is typically much smaller than both the hidden dimension \( d \) and the total number of nodes \( n \). Consequently, this \( O(n \cdot d \cdot f) \) term can be ignored in the asymptotic analysis.

Thus, by omitting the \( |TI| \) factor from the simultaneous encoder and the \( O(n \cdot d \cdot f) \) term for the decoder module, it can be concluded that the policy model proposed by Kool et al. and SED2AM have the same computational complexity in the context of asymptotic analysis.

\begin{table}[h!]
\centering

\caption{Computational Complexity of Each Component in Policy Networks}
\begin{tabular}{|c|c|c|}
\hline
\textbf{Component} & \textbf{Kool et. al Model \cite{kool2018attention}} & \textbf{SED2AM} \\
\hline
Encoder's First Sub-layer &\( O(n^2 \cdot d \cdot L) \) & \( O(n^2 \cdot d  \cdot L \cdot  |TI|) \) \\
\hline
Encoder's Second Sub-layer & \( O(n \cdot d^2 \cdot L) \) & \( O(n \cdot d^2 \cdot L \cdot |TI|) \) \\
\hline

Decoder Module (Per Decoding Step) & \( O(n \cdot d^2 + n^2 \cdot d) \) & \( O(n \cdot d^2 + n^2 \cdot d + n \cdot d \cdot f) \) \\

\hline
\end{tabular}
\label{tab:complexity}
\end{table}

\begin{table}[h!]
\centering
\caption{Description of Notations Used in Complexity Table}
\begin{tabular}{|c|c|}
\hline
\textbf{Notation} & \textbf{Description} \\
\hline
\( n \) & The number of customers plus the depot \\
\hline
\( d \) &  The dimension of the embeddings \\
\hline
\( L \) & Number of attention layers \\
\hline
\( |TI| \) & Number of time intervals\\
\hline
\( f \) & Number of vehicles in the fleet \\
\hline
\end{tabular}
\label{tab:complexity_notations}
\end{table}

\subsection{Policy Model Training} \label{training}

To train the policy model, a policy-gradient-based RL algorithm, named REINFORCE \cite{williams1992simple}, with a greedy rollout baseline is adopted. Given an initial state $s_0$ for a problem instance, the policy model is defined as a probability distribution $p_\theta(\Pi|s_0)$ parameterized by $\theta$. At the end of each episode, a solution $\Pi$ is attained by sequentially performing the vehicle selection and route construction decoding through sampling from the distribution $p_\theta(\Pi|s_0)$ at each decoding step. Given the parameter $\theta$, the loss $\mathcal{L}(\theta|s_0)$ is defined as follows:
\begin{eqnarray} \label{eq:prob}
\mathcal{L}(\theta|s_0) = \mathbb{E}_{p_\theta(\Pi|s_0)}\left[-R(\Pi)\right]
\end{eqnarray} 
Given the policy $p_\theta(\Pi|s_0)$, a baseline estimator of loss $b(\Pi)$, the gradient of loss $\nabla\mathcal{L}(\theta|s_0)$ is defined as:
\begin{eqnarray} \label{eq:loss}
\nabla\mathcal{L}(\theta|s_0) = \mathbb{E}_{p_\theta(\Pi|s_0)}\left[\left(-R(\Pi)-b(\Pi)\right)\nabla \log p_\theta(\Pi|s_0)\right]
\end{eqnarray} 
The greedy baseline utilized is a network parameterized by $\theta^{BL}$ with a similar neural structure as the policy model, but instead of sampling from the probability distributions presented by the vehicle selection decoder and trip construction decoder, it selects the vehicle and the node with the highest probability at each decoding step. This baseline is used for gradient variance reduction and faster convergence. 

The choice of the RL algorithm, specifically REINFORCE with a greedy rollout baseline, stems from its simplicity, computational efficiency that enhances scalability, and suitability for episodic combinatorial problems with sparse rewards, such as the MTTDVRP \cite{ahmadian2024back, anthony2017thinking, berto2023rl4co, zhang2021sample}.  The greedy rollout baseline enhances the REINFORCE algorithm by reducing gradient variance and improving stability during training, ensuring robust performance in complex decision-making tasks without introducing significant computational or implementation overhead \cite{rennie2017self}.

\section{Experiments} \label{section:experiments}
In this section, a comprehensive performance evaluation of the SED2AM model is presented, including comparisons with baseline models, an assessment of generalizability, sensitivity analysis, ablation studies, and an analysis of routing quality.

\subsection{Problem Sets Generation}
To evaluate the performance of the proposed SED2AM model, four sets of routing problems are generated, each containing a different number of customers (10, 20, 50, and 100). These problem sets are based on real truck speed data from two major Canadian cities, Edmonton and Calgary, with detailed characteristics listed in Table~\ref{tbl:problem_set}. For constructing the road network, OpenStreetMap \cite{boeing2017osmnx} is used to extract relevant truck routes within each city. Depot and customer node locations are uniformly sampled within city limits for each problem instance to ensure an even distribution.

To estimate travel times between locations, we implemented a time-dependent model that accounts for variations in truck speeds throughout the day. A step-function distribution is used to model the travel time for each edge \( e_{ij} \). The working hours for vehicles are divided into 10 equal time intervals (\( |TI| = 10 \)), each lasting 1.2 hours. For each interval, the travel speed on each road segment is estimated using averaging over truck speed data for the corresponding time interval, along with the estimated travel distance obtained from OpenStreetMap.

In Calgary, this speed data is derived from GPS records of truck movements collected in 2019, with data from over 133,000 GPS records taken from a single working day. For Edmonton, truck GPS data from 2019 is used as well, selecting a high-traffic day with over 600,000 GPS records. By averaging travel speeds within the time intervals, time-dependent travel times between nodes are estimated.

Customer demands are randomly generated for each problem instance, assigned from a discrete set of values within the range \{1, \dots, 9\}. Additionally, vehicle capacities for different problem sizes are provided in Table~\ref{tbl:problem_set}.

\begin{table*}[!h]
    \caption{Training and Test Set Sizes, and Vehicles Maximum Capacity for Problems of Different Size}
    \begin{adjustbox}{width=1\textwidth}
    \label{tbl:problem_set}
    {\LARGE
    \begin{tabular}{c|c | c | c | c | c }
        \hline
        \textbf{Problem Set} & \textbf{Problem Size} & \textbf{Train Set Size} & \textbf{Test Set Size} & \textbf{Vehicles Maximum Capacity} & \textbf {Number of Vehicles} \\ \hline
        MTTDVRP-10 & 10 & 512000 & 10000 & 20 & 2\\ 
        MTTDVRP-20 & 20 & 512000 & 10000 & 30 & 3\\
        MTTDVRP-50 & 50 & 512000 & 1000 & 40 & 3\\ 
        MTTDVRP-100 & 100 & 512000 & 1000 & 50 & 5\\ \hline
    \end{tabular}
    }
    \end{adjustbox}
\end{table*}

\subsection{Experimental Settings}
For training and testing the DRL-based models, servers equipped with A100 GPUs are used. All other baseline method experiments are conducted on servers with 32 cores and Intel(R) Xeon(R) Gold 6240R CPUs. The hyperparameters for the different DRL-based methods are presented in Table~\ref{tbl:hyperparameters}. Notably, in all experiments, the hidden dimensions for fully connected networks are set to the "hidden layers dimension" specified in the table. Early stopping is employed during training, halting when validation loss shows no further improvement.

\begin{table*}[!ht]
    \caption{Hyperparameters for DRL-based methods, where \( d \) represents the hidden dimension for node embeddings. The corresponding module is indicated in parentheses before each hyperparameter.}
    \label{tbl:hyperparameters}
    \begin{adjustbox}{width=0.8\textwidth}
    \LARGE
    \begin{tabular}{|c|c|c|c|}
        \hline
        \textbf{Hyperparameters} & \textbf{AM} & \textbf{E-GAT} & \textbf{SED2AM} \\ \hline
        Node embedding dimension (Encoder) & 128 & 128 & 128 \\ \hline
        Edge embedding dimension (Encoder) & - & 64 & 128 \\ \hline
        Hidden layers dimension (Encoder \& Decoder) & 128 & 128 & 128 \\ \hline
        Number of attention heads (Encoder \& Decoder) & 8 & 8 & 8 \\ \hline
        Logit clipping ($C$ in Equation \ref{eq:kqv_comp_decoder2_tanh} for Decoder) & 10 & 10 & 10 \\ \hline
        Number of attention layers (Encoder) & 3 & 3 & 3 \\ \hline
        Adam's initial learning rate (RL Optimizer) & $10^{-4}$ & $10^{-3}$ & $10^{-4}$ \\ \hline
        Learning rate decay schedule (RL Optimizer) & - & $0.96^{\text{epoch}}$ & $0.995^{\text{iteration}}$ \\ \hline
        Maximum number of epochs (RL Optimizer) & 500 & 500 & 500 \\ \hline
        Batch size (RL Optimizer) & 256 & 256 & 256 \\ \hline
        Initial model parameters distribution (Encoder \& Decoder) & Uniform & Uniform & Uniform \\ \hline
        Parameter initialization range (Encoder \& Decoder) & $\left[\frac{-1}{\sqrt{d}},\frac{1}{\sqrt{d}}\right]$ & $\left[\frac{-1}{\sqrt{d}},\frac{1}{\sqrt{d}}\right]$ & $\left[\frac{-1}{\sqrt{d}},\frac{1}{\sqrt{d}}\right]$ \\ \hline
    \end{tabular}
    \end{adjustbox}
\end{table*}

The following two decoding strategies are employed in the experiments conducted with SED2AM, E-GAT, and AM:

\begin{enumerate}
    \item \textbf{Greedy}: At each decoding step, the vehicle selection and trip construction decoders are used to select the vehicle and node with the highest probability, respectively.

    \item \textbf{Sampling}: This strategy involves 1280 solutions being sampled from the probability distributions output by the vehicle selection and trip construction decoders, with the most efficient route among them being selected as the solution.
\end{enumerate}

\subsection{Baselines}

For comparison, we adopt the following heuristic-based and state-of-the-art DRL-based methods as baselines:

\begin{enumerate}
  \item Ant Colony Optimization (ACO): Donati et al. \cite{donati2008time} proposed an enhanced version of the ACO algorithm, introduced to solve the time-dependent vehicle routing problem (TDVRP) with time windows, aiming to minimize total travel time through a novel pheromone updating and probability calculations. However, this approach does not support multi-trip routing.
  
  To address this limitation, we incorporated the early vehicle change technique introduced by Han et al. \cite{han2022solving}. By merging these methodologies, we developed a comprehensive baseline for the MTTDVRP that accommodates both time-dependence and multi-trip routing within the constraints of maximum working hours.  As for the parameters, the number of ants is set to 100, and the pheromone evaporation rate to 0.1. The number of iterations used for problems of different sizes is summarized in Table \ref{tbl:aco_prams}. To reduce the computational time, the ants’ route construction is parallelized on multiple cores of a server.

\begin{table*}[!h]
    \caption{The Number of Iterations of Ant Colony Optimization for Different Problem Sets}
    \label{tbl:aco_prams}
    \begin{adjustbox}{width=0.35\textwidth}
    \LARGE
    \begin{tabular}{c|c}
        \toprule 
        \textbf{Problem Sets} & \textbf{Number of Iterations} \\
        \hline 
        MTTDVRP-10 & 1280 \\
        MTTDVRP-20 & 2496 \\
        MTTDVRP-50 & 4992 \\
        MTTDVRP-100 & 12496 \\
        \bottomrule 
    \end{tabular}
    \end{adjustbox}
\end{table*}

  \item Genetic Algorithm (GA): For the GA baseline, we employ an elitist-preserving strategy, as used in the multi-vehicle trip generation method employed by Anggodo et al. \cite{anggodo2017optimization}. In this context, a value of $0$ signifies the separation between trips, while a gene value of $n+1$ marks the final trip for each vehicle. However, the approach used by Anggodo et al. does not consider time-dependent travel time. Hence, we adopted the fitness function by Jung et al. \cite{jung2001genetic} for our GA-based baseline to solve the MTTDVRP with a maximum working hours constraints. The fitness function, tailored for this VRP variant, is defined as follows:
\begin{eqnarray} \label{eq:fitness}
F(\Pi|S) = \frac{1}{1 - R(\Pi)} + \sum_{j \in \{1, \dots, f\}} (T_{max} + R(\Pi_{j})).
\end{eqnarray} 
where the reward function \( R \) is defined in Equation \ref{form:reward}, and \( R(\Pi_{j}) \) represents the total travel time for the set of trips performed by vehicle \( j \). For all problem sizes, the mutation rate and crossover rate are set to 0.1 and 0.9, respectively. Table \ref{tbl:ga_prams} summarizes the population size and maximum generation settings. To enhance algorithms' running time, the procedure is parallelized across multiple server cores.

  \item AM \cite{kool2018attention}: AM is a state-of-the-art DRL-based method with a Transformer-style encoder-decoder structure, proposed for solving several variants of TSP and VRP; however, this approach does not address multi-trip VRP. 
  To ensure adherence to maximum working hours, a masking condition is added. The experimental settings for this DRL-based method are summarized in Table \ref{tbl:hyperparameters}.
  \item E-GAT \cite{lei2022solve}: This is a DRL-based approach that exploits the edge information in the graph structure as well as node-related features. However, this approach is not suited for MTTDVRP. For a fair comparison, the encoding for each time-interval is separately performed. At each decoding step, the embeddings corresponding to that time-interval are used. 
  The experimental settings for this DRL-based method are summarized in Table \ref{tbl:hyperparameters}.

\end{enumerate}
It is noteworthy that the AM and E-GAT construct routes independently and do not associate the routes with specific vehicles or keep track of the remaining working hours of the vehicles. Since travel time is dynamic, merging routes and forming vehicle trips becomes ineffective. 

A possible approach involves constructing all routes for a single vehicle in a day as follows: start from the depot, serve customers until the vehicle’s capacity is reached, return to the depot to reload, and then continue serving more customers. This process repeats until no additional trips can be made, ensuring the vehicle returns to the depot within the maximum working hours constraints. However, additional returns to the depot due to the inability to fully utilize the vehicle’s capacity can make this method inefficient.

To promote fairer comparison, we explored two alternative strategies:
One, choosing the vehicle with the highest remaining working hours at each decoding step; and two, randomly picking a vehicle at each decoding step. The strategy of random vehicle selection yielded better overall routing performance, leading us to adopt this approach for our experiments.
\begin{table*}[!hb]
    \caption{Population Size and Maximum Generations for the Genetic Algorithm}
    \label{tbl:ga_prams}
    \begin{adjustbox}{width=0.5\textwidth}
    \LARGE
    \begin{tabular}{c|c c}
        \toprule 
        \textbf{Problem Sets} & \textbf{Population} & \textbf{Maximum Generations} \\
        \hline 
        MTTDVRP-10 & 128 & 200 \\
        MTTDVRP-20 & 256 & 500 \\
        MTTDVRP-50 & 1024 & 2000 \\
        \bottomrule 
    \end{tabular}
    \end{adjustbox}
\end{table*}

\subsection{Evaluation Metrics}
In this study, the performance is measured via the total traveling time, running time, as well as optimality gap in total traveling time. Assuming that $R(\Pi^{best})$ denotes the best result achieved, for a given solution $\Pi$, the optimality gap of that solution $Gap$ is defined as follows:

\begin{eqnarray} \label{eq:metrics}
Gap =  \frac{R(\Pi) - R(\Pi^{best})}{R(\Pi^{best})} \times 100\%.
\end{eqnarray} 
\subsection{Performance Analysis}

In this section, we aim to compare SED2AM with baseline methods, encompassing both traditional heuristic approaches and state-of-the-art DRL-based methods, across problems of four different sizes. The outcomes of the comparison with respect to the average objective value, its optimality gap, and the time consumed for solving each problem instance are summarized in Table \ref{tbl:performance_analysis1}. Due to the large computational time of the genetic algorithm, we are unable to provide the results of the GA for solving instances with 100 customers within a reasonable time.

As shown in Table \ref{tbl:performance_analysis1}, for the Calgary dataset, SED2AM with \(greedy\)-based decoding (SED2AM (Greedy)) surpasses two DRL-based baselines using the same decoding strategy across all problem sizes. The gap of improvement expands as the problem size increases. The expansion of the performance gap further highlights the SED2AM policy model's effectiveness in multi-trip time-dependent routing in the presence of maximum working hours constraints. Specifically, for problems larger than 50, SED2AM(Greedy) surpasses AM(Sampling) in performance, achieving an objective value of 1530.8 compared to 1466.6. Moreover, in scenarios involving 100 customers, it outperforms GAT(Sampling) by reducing the optimality gap from 10.75\% to 3.77\%.

Compared to heuristic methods, SED2AM(Greedy) exhibits a slightly higher objective value than ACO for problem sizes of 10 and 20, but it surpasses the performance of this heuristic method for problem sizes of 50 and 100. Additionally, SED2AM(Greedy) requires significantly less computational time compared to this meta-heuristic approach. The difference in running time between SED2AM(Greedy) and GA is much greater than that with ACO, despite the slightly higher objective value. SED2AM(Sampling), in comparison, outperforms both meta-heuristic methods in terms of both running time and routing performance. Although employing a \(sampling\)-based strategy for DRL methods leads to longer run times, they significantly outperform their \(greedy\)-based counterparts in terms of the objective. SED2AM, with either decoding strategy, exhibits better performance compared to the DRL-based methods using the same strategies.

The experiments conducted on the Edmonton dataset yield similar outcomes for SED2AM, with one notable exception: for problems of size 50, SED2AM(Greedy) slightly outperforms E-GAT(Sampling).  Overall, the experimental results demonstrate that SED2AM outperforms the baselines in solving MTTDVRP with maximum working hours.

\begin{table*}[!htbp]
    \caption{SED2AM vs. the baselines for solving different size routing problems in Calgary and Edmonton, Canada. Objective value (Obj.) represents total travel time, while the Gap indicates the optimality gap, reflecting the normalized deviation from the optimal result. The time column for heuristic methods (ACO, GA) refers to the average problem-solving duration, whereas for DRL-based methods (AM, GAT, SED2AM), it denotes the average inference time. MTTDVRP-X specifies the problem with X customers. Greedy and Sampling respectively denote \(greedy\)-based and \(sampling\)-based decoding. The \(\downarrow\) indicates that a lower value is better.}
    \label{tbl:performance_analysis1}
    \begin{adjustbox}{width=1\textwidth}
    \Huge
    \begin{tabular}{l | c c c | c c c | c c c | c c c }
        \toprule
        & \multicolumn{3}{c|}{\textbf{MTTDVRP-10}} & \multicolumn{3}{c|}{\textbf{MTTDVRP-20}} & \multicolumn{3}{c|}{\textbf{MTTDVRP-50}} & \multicolumn{3}{c}{\textbf{MTTDVRP-100}} \\
        
        \textbf{Method} & Obj. $\downarrow$ & Gap $\downarrow$ & Time $\downarrow$ & Obj. $\downarrow$ & Gap $\downarrow$ & Time $\downarrow$ & Obj. $\downarrow$ & Gap $\downarrow$ & Time $\downarrow$ & Obj. $\downarrow$ & Gap $\downarrow$ & Time $\downarrow$ \\
        \hline
        
        & \multicolumn{11}{c}{\textbf{Calgary}} \\
        \cline{2-13}
        ACO & 463.4 & 0.22\% & 8.5 mins & 767.6 & 2.36\% & 35.9 mins & 1492.5 & 8.16\% & 2.6 hours & 3108.1 & 11.48\% & 13.5 hours \\
        GA & 464.4 & 0.78\% & 9.5 mins & 751.2 & 1.46\% & 42.8 mins & 1439.3 & 4.30\% & 12.1 hours & - & - & - \\
        AM(Greedy) & 486.1 & 5.49\% & 0.13 s & 799.3 & 6.59\% & 0.25 s & 1547.8 & 12.17\% & 0.57 s & 3328.1 & 19.38\% & 1.1 s \\
        AM(Sampling) & 481.0 & 4.38\% & 73.8 s & 778.1 & 3.76\% & 2.9 mins & 1530.8 & 10.94\% & 7.1 mins & 3163.3 & 13.46\% & 14.8 mins \\
        E-GAT(Greedy) & 483.7 & 4.99\% & 0.29 s & 788.2 & 5.11\% & 0.52 s & 1501.8 & 8.83\% & 1.39 s & 3254.3 & 16.72\% & 2.75 s \\
        E-GAT(Sampling) & 471.9 & 2.28\% & 2.3 mins & 779.5 & 3.95\% & 6.1 mins & 1465.2 & 6.18\% & 13.5 mins & 3087.7 & 10.75\% & 27.6 mins \\
        SED2AM(Greedy) (ours) & 472.0 & 2.43\% & 0.51 s & 793.9 & 5.87\% & 1.37 s & 1466.6 & 6.28\% & 3.49 s & 2893.2 & 3.77\% & 7.88 s \\
        SED2AM(Sampling) (ours) & \textbf{460.8} & 0.00\% & 4.5 mins & \textbf{749.9} & 0.00\% & 11.2 mins & \textbf{1379.9} & 0.00\% & 24.6 mins & \textbf{2787.9} & 0.00\% & 51.3 mins \\
        
        \hline
        & \multicolumn{11}{c}{\textbf{Edmonton}} \\
        \cline{2-13}
        ACO & 317.2 & 0.35\% & 8.2 mins & 490.7 & 3.28\% & 31.4 mins & 1083.9 & 12.07\% & 2.3 hours & 2229.2 & 13.03\% & 12.7 hours \\
        GA & 313.8 & 0.44\% & 9.1 mins & 481.3 & 1.30\% & 38 mins & 1006.1 & 4.03\% & 11.6 hours & - & - & - \\
        AM(Greedy) & 328.7 & 5.20\% & 0.13 s & 509.8 & 7.30\% & 0.24 s & 1085.3 & 12.22\% & 0.55 s & 2384.5 & 20.91\% & 1.04 s \\
        AM(Sampling) & 323.9 & 3.74\% & 75.3 s & 496.5 & 4.50\% & 3.1 mins & 1070.4 & 10.1\% & 7.0 mins & 2266.7 & 14.93\% & 14.7 mins \\
        E-GAT(Greedy) & 325.1 & 4.03\% & 0.27 s & 502.9 & 5.85\% & 0.54 s & 1054.4 & 9.02\% & 1.37 s & 2321.6 & 17.72\% & 2.77 s \\
        E-GAT(Sampling) & 317.0 & 1.44\% & 2.3 mins & 497.5 & 4.71\% & 6.2 mins & 1024.8 & 5.96\% & 13.1 mins & 2182.2 & 10.65\% & 28.1 mins \\
        SED2AM(Greedy) (ours) & 318.0 & 1.79\% & 0.48 s & 497.9 & 4.79\% & 1.35 s & 1024.0 & 5.88\% & 3.38 s & 2067.3 & 4.82\% & 8.03 s \\
        SED2AM(Sampling) (ours) & \textbf{312.5} & 0.00\% & 4.3 mins & \textbf{475.1} & 0.00\% & 10.8 mins & \textbf{967.1} & 0.00\% & 24.3 mins & \textbf{1972.1} & 0.00\% & 53.7 mins \\
        
        \bottomrule
    \end{tabular}
    \end{adjustbox}
\end{table*}

\subsection{Generalization on Problems of Different Scales}

Considering the infeasibility of training the model on large instances due to computational and resource constraints, investigating the generalizability of the model becomes a critical concern. In this experiment, the generalizability of SED2AM is investigated by applying the policy model trained on instances of a certain problem size to instances with a larger number of customers by conducting experiments using the Edmonton dataset, as this form of generalizability is important for real-world applications.

\begin{figure}[h!]
    \includegraphics[width=5in]{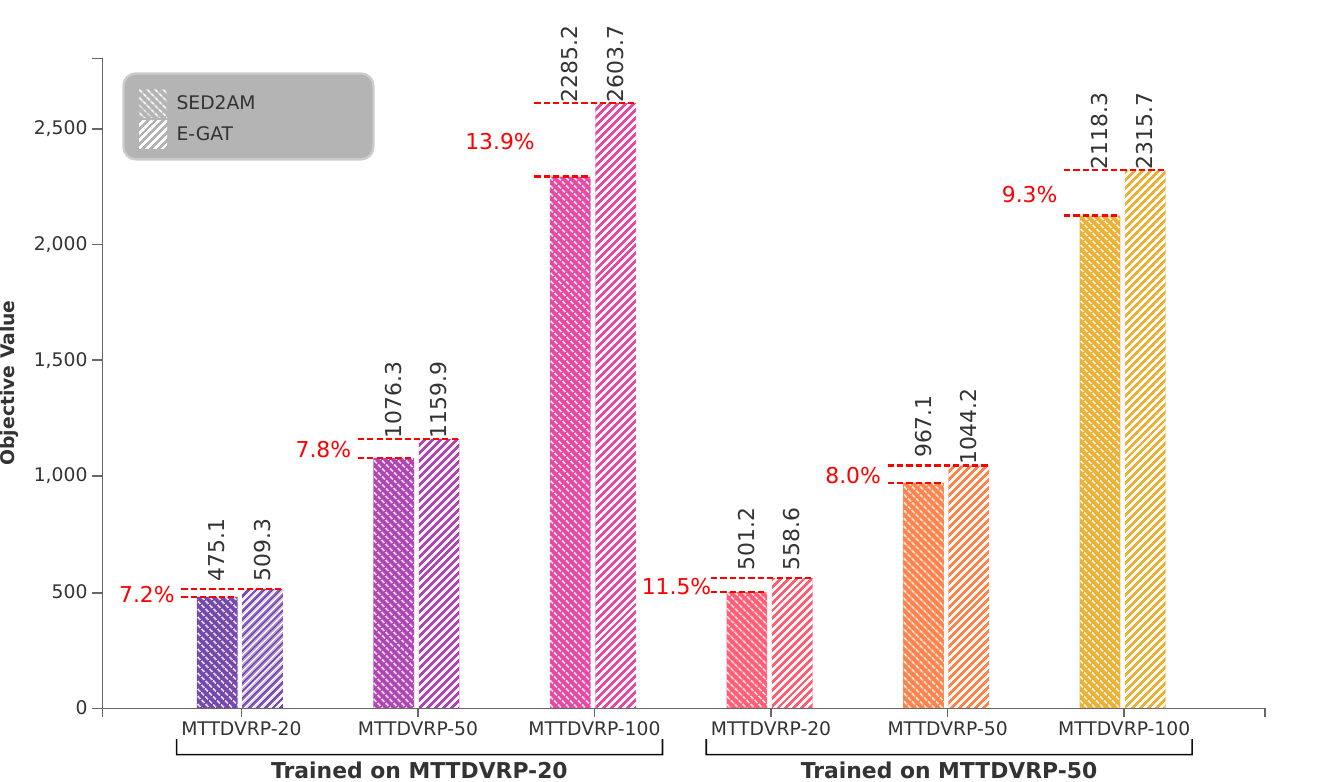}
  
  \caption{
This image compares the generalization capabilities of SED2AM and E-GAT across various problem sizes, showcasing policies that were trained and tested on differing scales. On the X-axis, "MTTDVRP-X" represents the evaluation of models on problems of size X, where each model is denoted as "Trained on MTTDVRP-Y," indicating it was initially trained on instances of size Y.
  }
  \label{diagram:generalization}
\end{figure}

Figure \ref{diagram:generalization} illustrates the performance results of E-GAT and SED2AM on MTTDVRP problem instances with 20 and 50 customers when tested across MTTDVRP-20, MTTDVRP-50, and MTTDVRP-100 scenarios. As shown in the figure, better performance is engendered when the policy trained on instances of a certain problem size is tested on problems of the same size rather than the other sizes.

It can also be observed that SED2AM achieves better performance in comparison to E-GAT, particularly when the learned policy is implemented across problem instances of varying sizes. Notably, the performance gap between E-GAT and SED2AM, favoring the latter, widens as the sizes of the test problem instances increase.

Moreover, while training the SED2AM model on instances of size 20 and testing it on MTTDVRP-100 leads to an increase in the objective value compared to when both training and testing are done on MTTDVRP-100 instances, it still outperforms ACO and AM(Greedy). Notably, when SED2AM is trained on MTTDVRP-50 and tested on MTTDVRP-100, it outperforms all baseline models except for E-GAT(Sampling), with which it performs comparably.

Overall, the experimental results summarized in Figure \ref{diagram:generalization} demonstrate the generalizability of the SED2AM approach.

\subsection{Comparative Performance Analysis with Varied Maximum Working Hours Constraints}

The maximum working hours are inherently shaped by regulatory constraints on working hours, which differ across countries. For instance, while Canadian regulations allow for a 13-hour maximum, the limits are 12 hours in Australia, 9 hours in the UK, and 10 hours in the US. To this end, we conducted additional experiments investigating the efficacy of our proposed solution for the multi-trip time-dependent vehicle routing problem under varying constraints of maximum working hours and comparing it with that of ACO, AM, and E-GAT. In this experiment, we set the number of vehicles to 5 and compared the routing performance of SED2AM with the sampling-based decoding strategy with that of E-GAT(Sampling), AM(Sampling), and ACO for solving MTTDVRP-100 and summarized the results in Table \ref{tbl:varyhours}. As perceivable from the table, SED2AM(Sampling), our method, outperforms all baselines. It can also be observed that as the maximum working hours constraints decreases, SED2AM exceeds other DRL-based baselines' performance by a larger gap, manifesting the power of SED2AM's policy network in exploiting the remaining working hours of the vehicles for better decision-making.

\begin{table*}[t]
    \caption{SED2AM vs. the Baselines (ACO as a heuristic method, and AM and GAT as DRL-based ones) for various maximum working hours constraints (6, 8, 10, and 12 hours) on both datasets (Calgary and Edmonton, Canada). \(Sampling\)-based strategy is adapted for the DRL-based methods. Objective value (Obj.), with a lower number indicating better performance, represents total travel time, while the Gap indicates the optimality gap, reflecting the normalized deviation from the optimal result. The \(\downarrow\) indicates that a lower value is better.}
    \label{tbl:varyhours}
    \begin{adjustbox}{width=.9\textwidth}
    \large
    \begin{tabular}{l | c | c c | c c | c c | c c}
        \toprule 
        &  & \multicolumn{2}{c}{{6 Hrs}} & \multicolumn{2}{c}{{8 Hrs}} & \multicolumn{2}{c}{{10 Hrs}} & \multicolumn{2}{c}{{12 Hrs}} \\
        
        & \multicolumn{1}{c|}{Method} & Obj. $\downarrow$ & Gap $\downarrow$ & Obj. $\downarrow$ & Gap $\downarrow$ & Obj. $\downarrow$ & Gap $\downarrow$ & Obj. $\downarrow$ & Gap $\downarrow$ \\
        
        \hline 
        \parbox[t]{2mm}{\multirow{4}{*}{\rotatebox[origin=c]{90}{\textbf{Calgary}}}} & ACO & 3127.5 & 11.76\% & 3121.4 & 11.80\% & 3113.5 & 11.70\% & 3108.1 & 11.48\% \\
        &AM(Sampling) & 3265.9 & 16.71\% & 3199.6 & 14.76\% & 3162.8 & 13.47\% & 3163.3 & 13.46\% \\
        &E-GAT(Sampling) & 3143.6 & 12.34\% & 3100.7 & 11.96\% & 3088.1 & 10.79\% & 3087.7 & 10.75\% \\
        &SED2AM(Sampling) & \textbf{2798.3} & 0.00\% & \textbf{2789.1} & 0.00\% & \textbf{2788.5} & 0.00\% & \textbf{2787.9} & 0.00\% \\
        
        \hline 
        \parbox[t]{2mm}{\multirow{4}{*}{\rotatebox[origin=c]{90}{\textbf{Edmonton}}}} & ACO & 2250.0 & 13.79\% & 2242.4 & 13.59\% & 2232.8 & 13.19\% & 2229.2 & 13.03\% \\
        &AM(Sampling) & 2371.5 & 19.93\% & 2303.9 & 16.70\% & 2275.3 & 15.35\% & 2266.7 & 14.93\% \\
        &E-GAT(Sampling) & 2243.9 & 13.48\% & 2209.2 & 11.90\% & 2185.8 & 10.81\% & 2182.2 & 10.65\% \\
        &SED2AM(Sampling) & \textbf{1977.3} & 0.00\% & \textbf{1974.1} & 0.00\% & \textbf{1972.5} & 0.00\% & \textbf{1972.1} & 0.00\% \\
        
        \bottomrule 
    \end{tabular}
    \end{adjustbox}
\end{table*}

\subsection{Ablation Study}
To solve the MTTDVRP with the maximum working hours constraints, two major improvements to AM, namely, the simultaneous encoder and vehicle selection decoder, are introduced. To investigate the influence of each, an ablation study on MTTDVRP-50 and MTTDVRP-100 is conducted on the Edmonton dataset, and the results are summarized in Table \ref{tbl:ablation}, which includes the average objective value, and the computational time of an instance. SED2AM(Greedy) without SE denotes that the simultaneous encoder is removed from the policy model's architecture of SED2AM, and results are collected through the \(greedy\) decoding strategy. VSD corresponds to the vehicle selection decoder module while SE stands for the simultaneous encoder. It is observable that both modules contribute to improving the routing performance; we attest to the rationale of the policy network design. Although the addition of the simultaneous encoder results in longer computational times, it contributed more to improving the routing performance than the vehicle selection decoder, which introduces a much shorter addition to the inference time. The lower objective value obtained by solely replacing AM's encoder with the proposed simultaneous encoder compared to that of E-GAT, as recorded in Table \ref{tbl:performance_analysis1}, demonstrates that SE is able to effectively capture time-dependent traveling time information. Additionally, the better performance of SED2AM(Greedy) without SE compared to AM demonstrates the addition of the vehicle selection decoder, and capturing the status of the fleet empowers the policy model to make more rational decisions in the view of multi-trip routing with maximum working hours. It is noteworthy that for SED2AM(Greedy) without SE, the vehicle selection decoder is not supplied with time-dependent contextual information. Similarly, for the case of SED2AM(Greedy) without VSD, the route selection decoder utilizes the remaining capacity of the vehicle instead of a linear projection of the currently selected vehicle's state.

\begin{table*}[h!]
    \caption{Ablation study results for major modifications to the policy model. VSD represents the vehicle selection decoder, and SE refers to the simultaneous encoder. The objective value (Obj.) is the average total travel time, with lower values indicating improved routing performance. Time denotes the corresponding method's average inference time. The $\downarrow$ indicates that a lower value is better.}
    \label{tbl:ablation}
    \begin{adjustbox}{width=.6\textwidth}
    \LARGE 
    \begin{tabular}{l | c c | c c}
        \toprule 
        & \multicolumn{2}{c|}{{MTTDVRP-50}} & \multicolumn{2}{c}{{MTTDVRP-100}} \\
        
        \multicolumn{1}{c|}{Method} & Obj. $\downarrow$ & Time $\downarrow$ & Obj. $\downarrow$ & Time $\downarrow$ \\
        \hline 
        SED2AM(Greedy) without VSD & 1046.9 & 4.09 s & 2091.0 & 9.79 s \\
        SED2AM(Greedy) without SE & 1079.5 & 1.58 s & 2244.7 & 3.94 s \\
        SED2AM(Greedy) & 1024.0 & 6.39 s & 2067.3 & 14.3 s \\
        \bottomrule
    \end{tabular}
    \end{adjustbox}
\end{table*}

\subsection{Comparative Performance Analysis with Different Vehicle Selection Strategies}
AM and E-GAT do not employ any method to directly associate the routes to the vehicles that execute them, they construct trips independently, without considering which specific vehicle is assigned. This approach has significant limitations in scenarios where vehicles are allowed to perform multiple trips. Specifically, there is no mechanism to ensure compliance with constraints such as maximum working hours or cumulative travel times for individual vehicles during route construction. 

To address this issue and employ a method that facilitates multi-trip route construction while ensuring compliance with maximum working hours constraints, we developed three different strategies. The first strategy is the single-vehicle strategy, where a vehicle constructs trips sequentially until its maximum working hours are fully utilized, ensuring it can return to the depot after serving the last customer within the maximum working hours constraints. The second strategy is the maximum remaining working hours strategy, which selects the vehicle with the most remaining working hours at each step of trip construction. The third strategy is the random vehicle selection strategy, where a vehicle is randomly chosen at each decoding step to construct the next trip.

In this section, we compare the performance of the three proposed methods with that of SED2AM, both with and without the vehicle selection decoder (using the random selection strategy). It is important to note that greedy decoding is employed for all experiments. The results, summarized in Table \ref{tbl:new_exp}, demonstrate that the best performance is achieved by SED2AM with the vehicle selection decoder, followed by SED2AM without the decoder, but utilizing the random vehicle selection strategy (SED2AM w/o VSD).

Notably, the random selection strategy (Random) used for AM and E-GAT throughout the paper outperforms both the maximum remaining working hours strategy (Max) and the single-vehicle strategy (Single). This highlights the effectiveness of random selection as the method employed for vehicle selection in the rest of the experiments for AM and E-GAT.

\begin{table*}[h!]
    \caption{Comparison of vehicle selection strategies at decoding steps: maximum remaining working hours (Max), random selection (Random), single vehicle routing until the maximum working hours are reached (Single), and SED2AM without vehicle selection decoding but with random selection (SED2AM w/o VSD), illustrating their impact on routing performance for MTTDVRP across different problem scales. A lower value ↓ indicates better performance.}
    \label{tbl:new_exp}
    \begin{adjustbox}{width=0.5\textwidth}
    \Huge
    \begin{tabular}{l | c | c}
        \toprule
        {\textbf{Method}} & 
        \multicolumn{1}{c|}{\textbf{MTTDVRP-50}} & 
        \multicolumn{1}{c}{\textbf{MTTDVRP-100}}\\

        \hline
        AM(Max) & 
        1089.5 & 2389.6   \\
        AM(Random) & 
        1085.3 & 2384.5  \\
        AM(Single) & 
        1094.8 & 2396.0   \\
        \hline
        E-GAT(Max) & 
        1059.5 & 2327.8   \\  
        E-GAT(Random) & 
        1054.4 &  2321.6  \\   
        E-GAT(Single) & 
        1063.9 & 2332.4    \\  
        \hline         
        SED2AM w/o VSD & 
        1046.9 & 2091.0     \\
        SED2AM & 
        1024.0 & 2067.3    \\
        
        \bottomrule
    \end{tabular}
    \end{adjustbox}
\end{table*}

\subsection{Routing Quality Analysis}

\begin{figure*}[!htbp]
\centering
\subfloat[7:00 AM]
{\label{fig:qualityama}
    \includegraphics[width=1\textwidth]{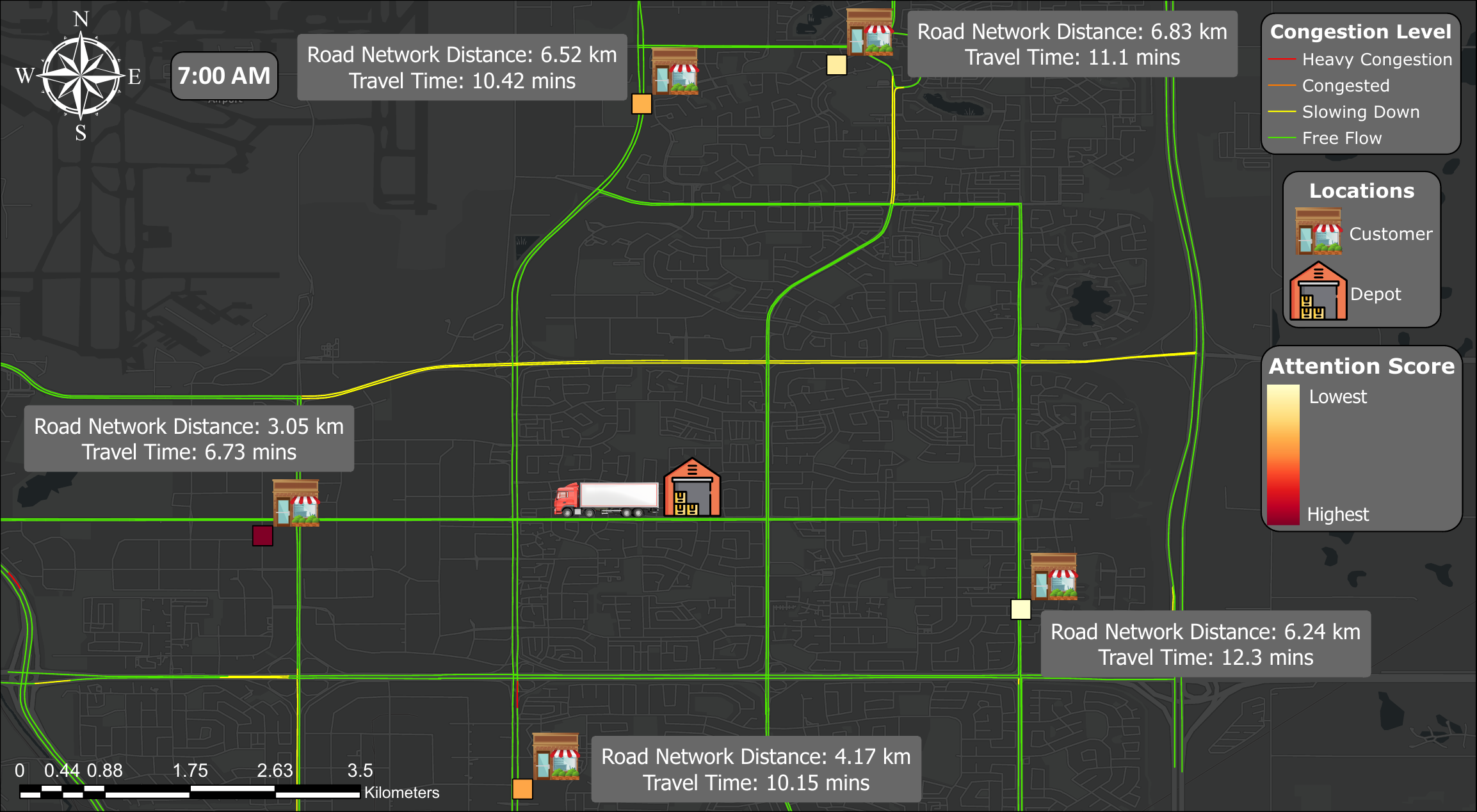}}

\subfloat[4:30 PM] {
\label{fig:qualityamb}
    \includegraphics[width=1\textwidth]{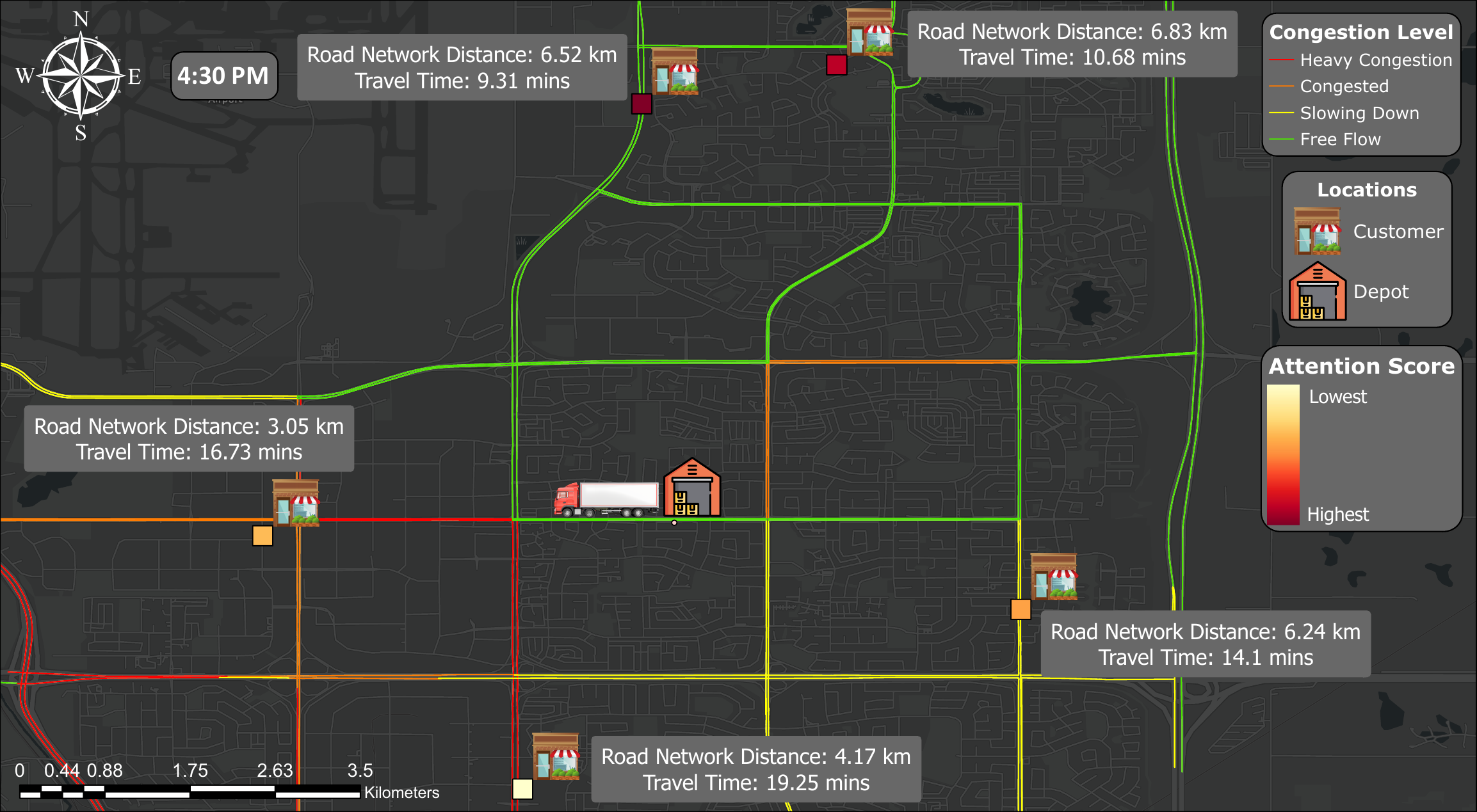}}
\caption{This figure illustrates the routing process for the Multi-Trip Time-Dependent Vehicle Routing Problem with 20 customers (MTTDVRP-20) based on a Calgary dataset instance, showcasing variations at two distinct start times: 7:00 AM and 4:30 PM. Adjacent to the customer locations, the vertical bars' color represents the softmax normalized attention scores for the 5 locations with the highest scores among the twenty customers. The violet path highlights the path from the depot to the customer with the highest attention score.
}
\label{fig:qualityam}

\end{figure*}

In this section, we aim to illustrate the capability of SED2AM in decision-making based on the time-dependent traveling time between the nodes. To achieve this, we present the softmax normalized attention scores calculated during the decoding steps.
To better understand the encoder's effectiveness without the influence of the VSD module on routing and the constraints on vehicles' maximum operating hours, we excluded the vehicle selection decoder and the masking condition related to vehicle operational time limits. The experiments reported here are conducted using data from the Calgary dataset.
In the first experiment, we explored the ability of SED2AM's simultaneous encoder to capture time-dependent travel times and to utilize it for the decoders' decision-making. For an MTTDVRP-20 instance, we compared the five nodes with the highest attention scores (softmax normalized among these nodes) at the first decoding step, for two different start times. The results are shown in Figure \ref{fig:qualityam}.
The map depicted in Figure \ref{fig:qualityama} shows that the traffic on most road segments on the road networks originating from the depot and leading to the five specified customer nodes is generally light (shown as green).
The boxes adjacent to the customer nodes display the road network travel distance and travel time from the depot to each node. The color of the vertical bars beside each customer node indicates the attention scores. At \textit{7:00 AM}, the attention scores reveal that the nodes located to the west and south-west of the depot are assigned higher attention scores. This indicates a greater likelihood of these nodes being selected at the current decoding step.
Conversely, the nodes situated to the north and east, characterized by longer travel times, are assigned lower attention scores.  At \textit{4:30 PM}, Figure \ref{fig:qualityamb} illustrates heavy traffic conditions on the routes from the depot to the nodes located to the west and south-west, leading to significantly increased travel times compared to those recorded at \textit{7:00 AM}, as well as compared to the travel times to nodes in the north and east at the current time. This example demonstrates that SED2AM's policy model effectively utilizes the travel times between the nodes for decision-making.

\begin{figure*}[!htbp]

\centering

\subfloat[8:00 AM]
{   
\label{fig:qualitygata}

\includegraphics[width=0.32\textwidth]{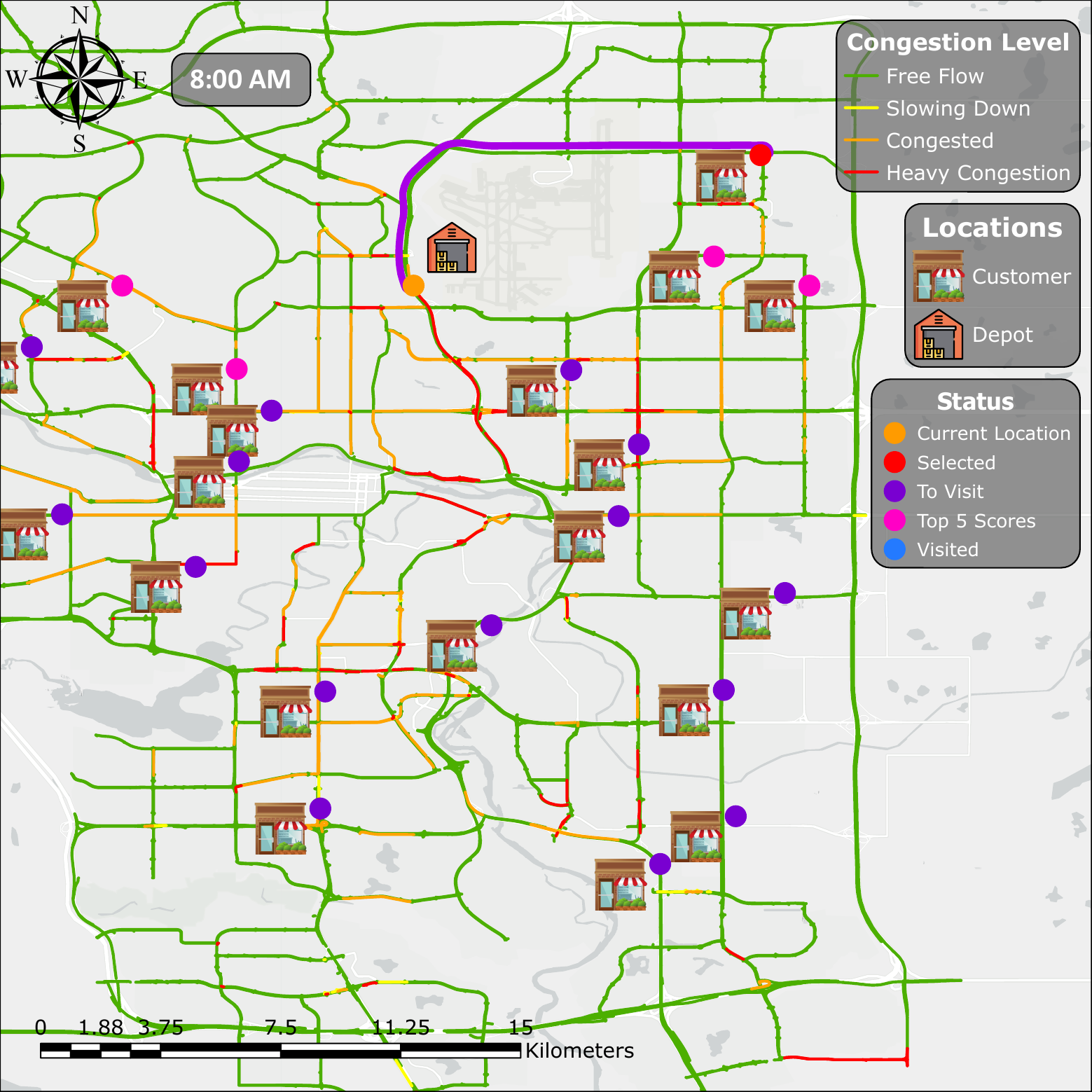}}
\hspace{-0.2cm}
\subfloat[8:35 AM]
    {
\label{fig:qualitygatb}
    \includegraphics[width=0.32\textwidth]{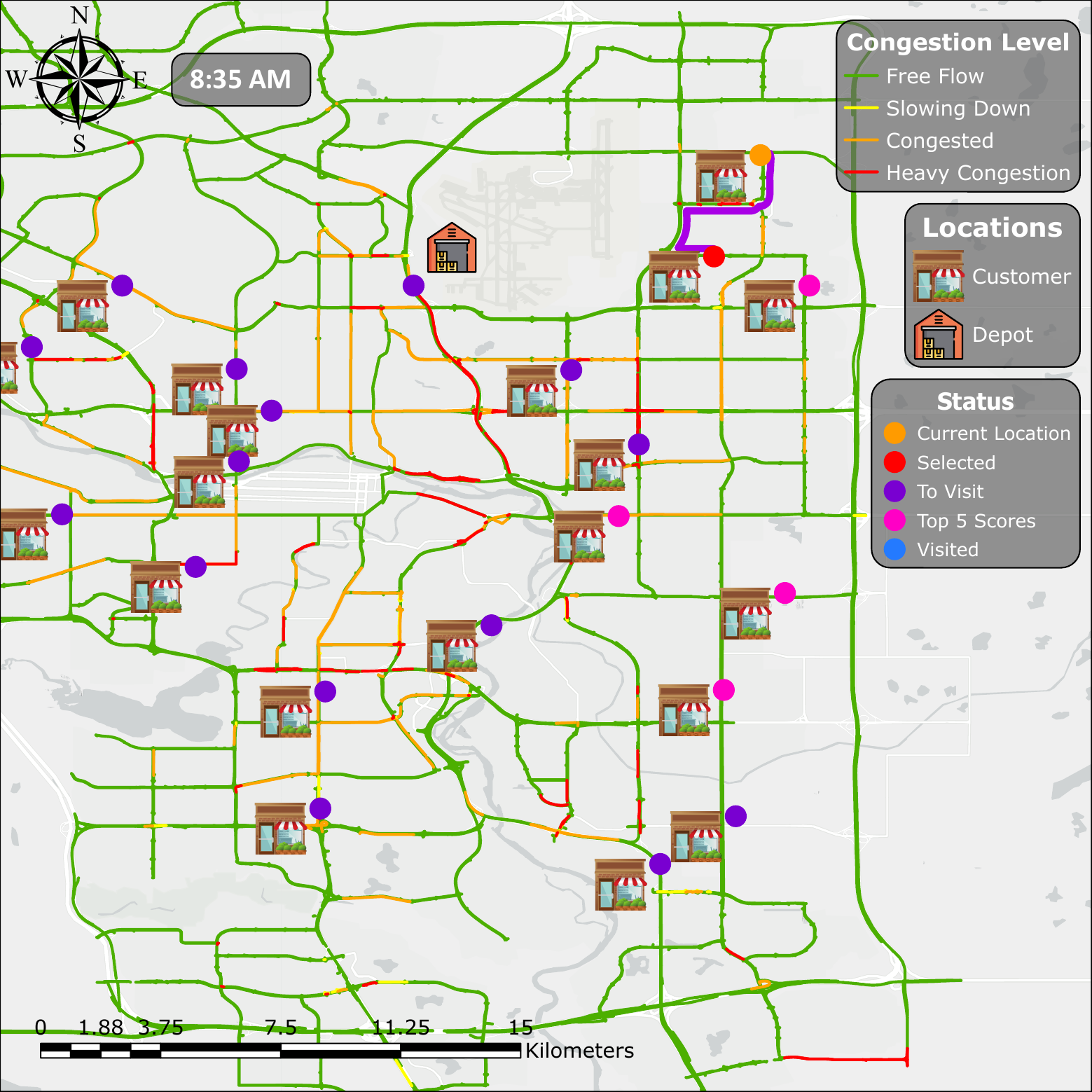}}
\hspace{-0.2cm}
\subfloat[9:07 AM]
    {
\label{fig:qualitygatc}
    
    \includegraphics[width=0.32\textwidth]{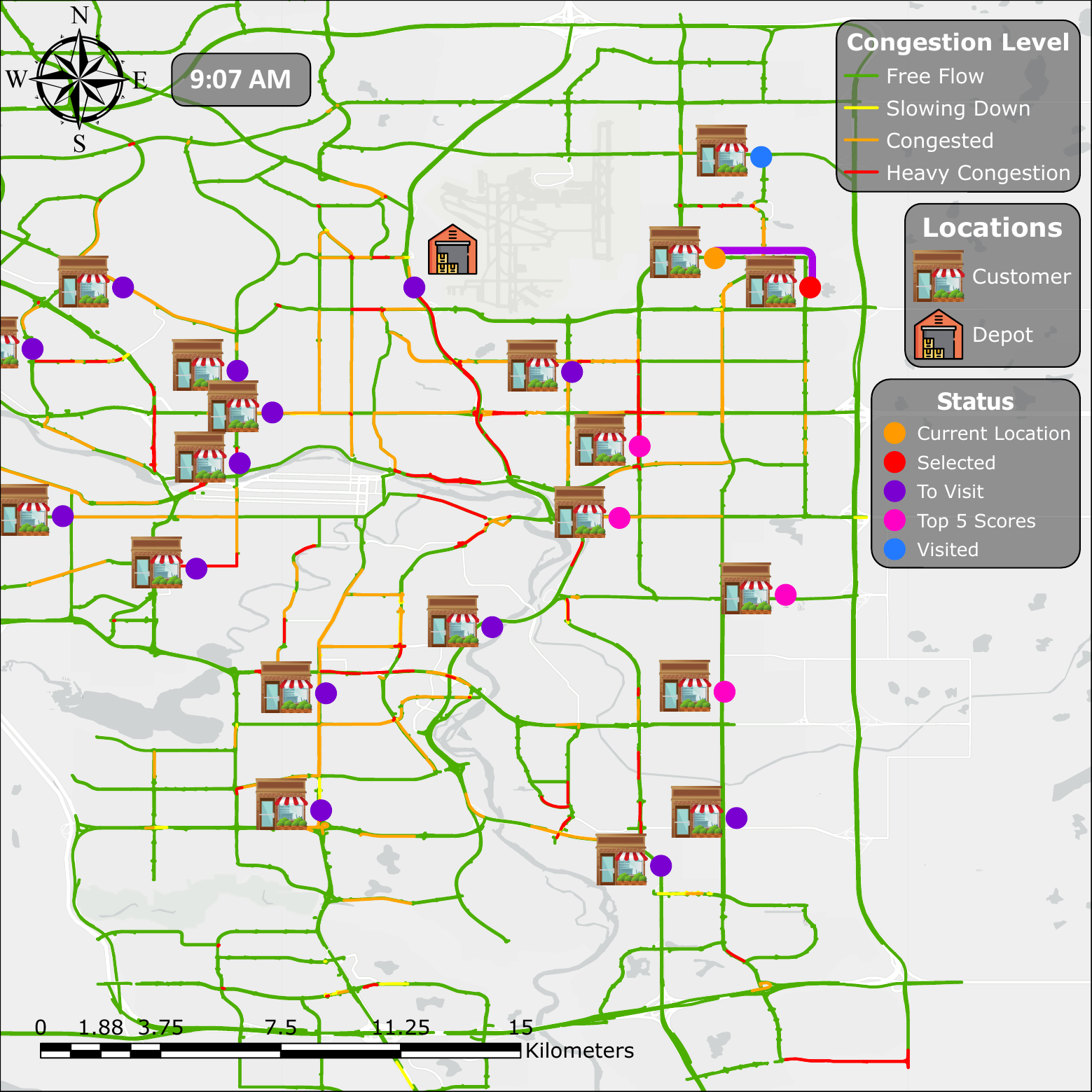}}
\vspace{-0.3cm}
\subfloat[9:18 AM]
    {
    
    \label{fig:qualitygatd}
    \includegraphics[width=0.32\textwidth]{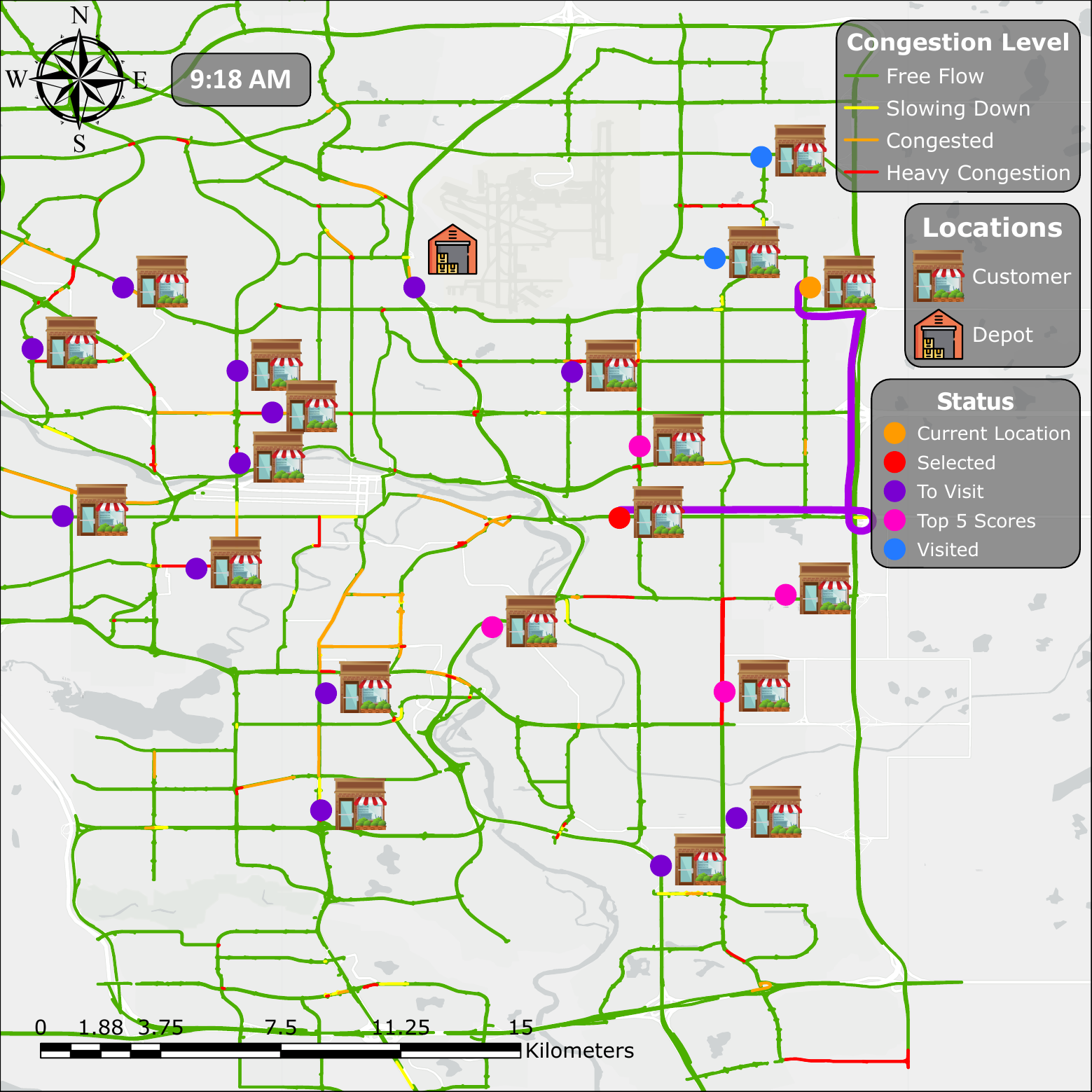}}
\hspace{-0.2cm}
\subfloat[9:42 AM]
    {
    \label{fig:qualitygate}
    \includegraphics[width=0.32\textwidth]{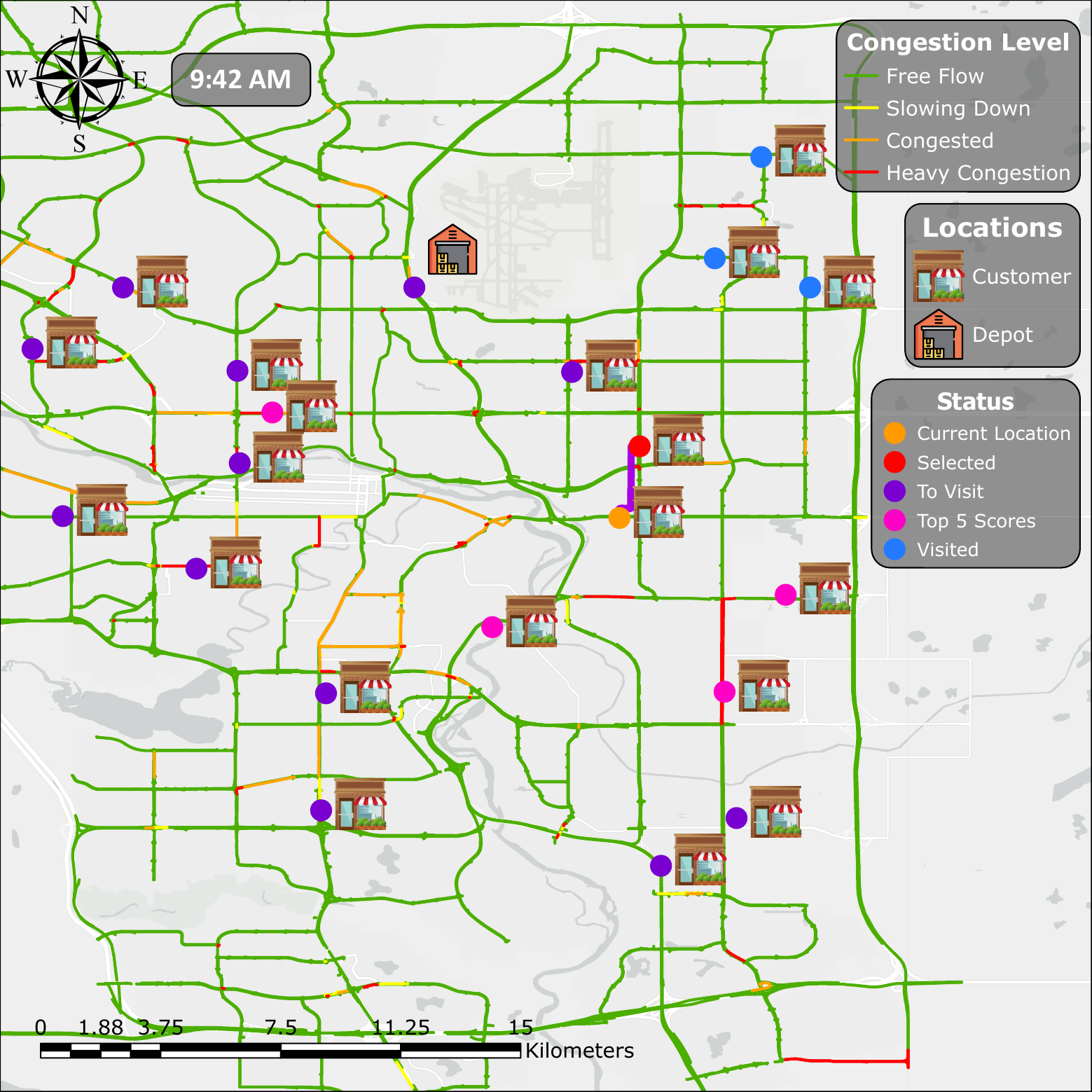}}
\hspace{-0.2cm}
\subfloat[9:50 AM]
{ 
\label{fig:qualitygatf}
\includegraphics[width=0.32\textwidth]{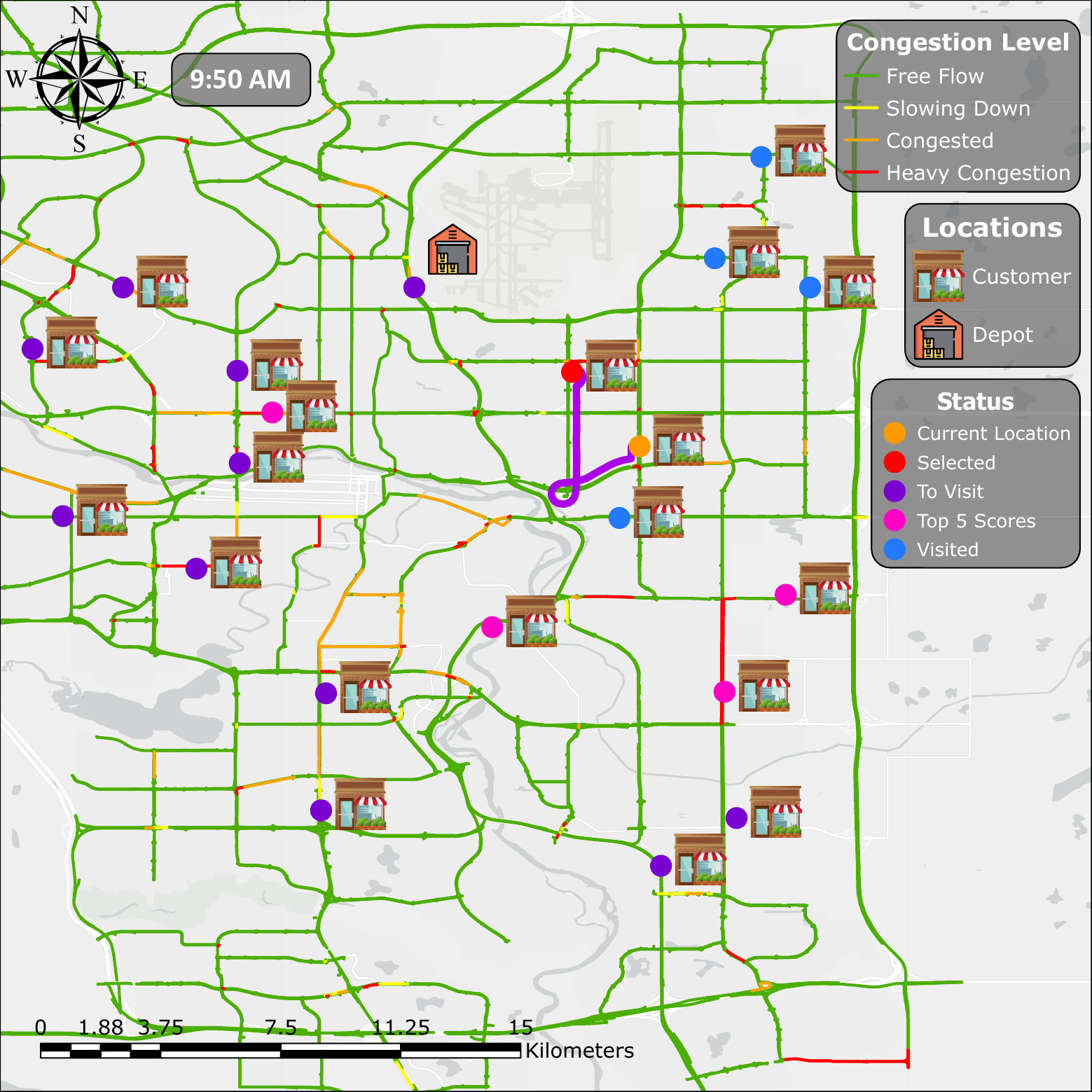}}
\vspace{-0.3cm}
\hspace{0.15cm}
\subfloat[10:01 AM]
    {
\label{fig:qualitygatg}
    \includegraphics[width=0.32\textwidth]{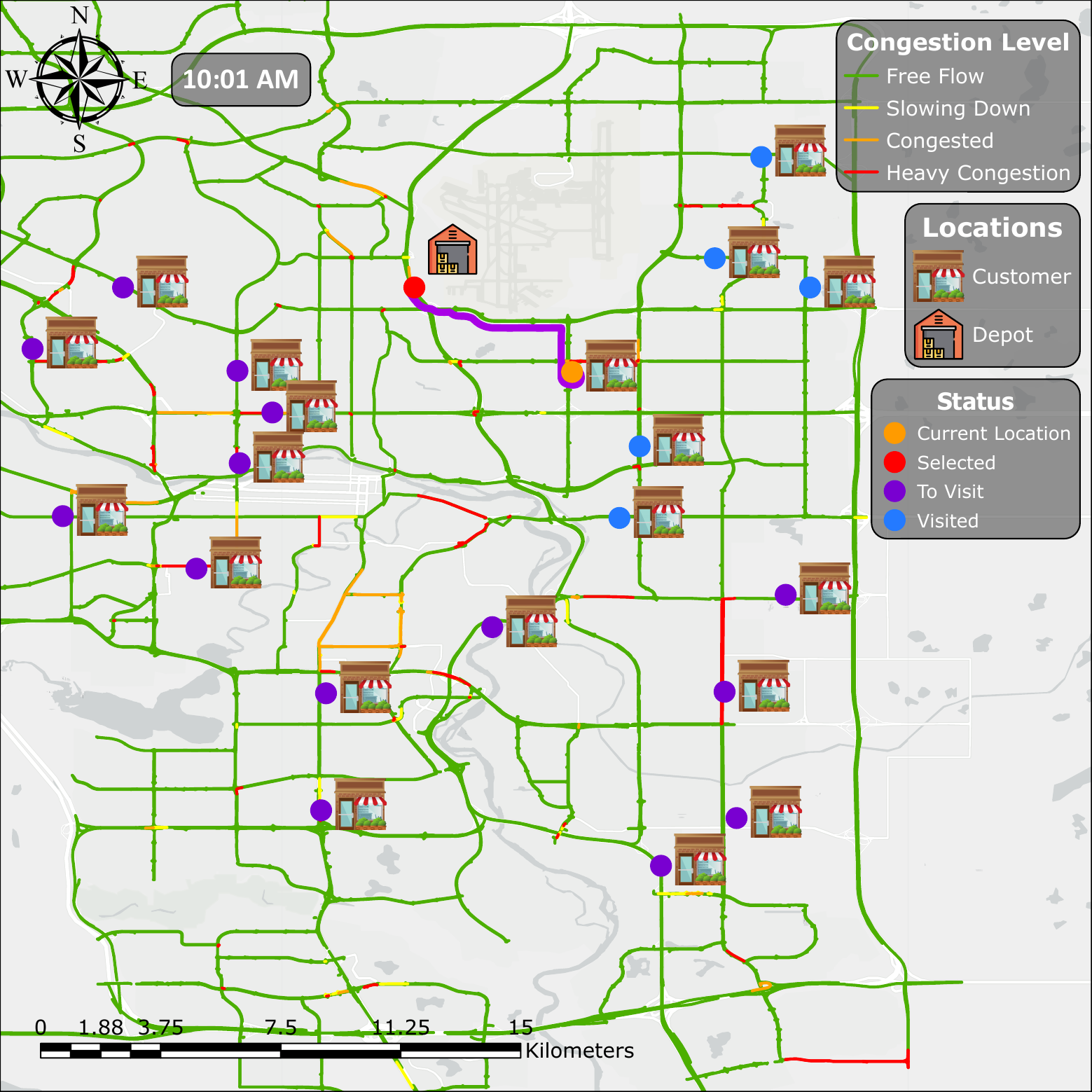}}
\caption{Routing Process for MTTDVRP-20 on a Calgary dataset instance. The path shown in violet indicates the route between the current location of the vehicle and the node to be visited in the subsequent decoding step.
}
\label{fig:qualitygat}
\end{figure*}

In a subsequent experiment, we demonstrated the route construction process for an MTTDVRP-20 instance starting at \textit{8:00 AM} under varying traffic conditions, as illustrated in Figure \ref{fig:qualitygat}. In this figure, at each decoding step, the node highlighted with a yellow halo represents the vehicle's current location. The node selected as the next destination at the decoding step is marked in red, while the other nodes within the top 5, having the highest attention scores, are indicated in violet. Starting from \textit{8:00 AM}, the vehicle departs the depot to serve a customer located far to the north. As shown in  Figure \ref{fig:qualitygata}, at this decoding step, SED2AM prioritizes nodes that are on paths with light traffic (indicated by green lines) over those in congested areas (marked by red lines), despite their proximity, during the morning rush hour. In the subsequent two decoding steps, as shown in Figure \ref{fig:qualitygatb} and \ref{fig:qualitygatc}, nodes near the initially selected node and situated on routes with free-flowing traffic are chosen. For the next decoding steps, presented in  Figure \ref{fig:qualitygatd}, \ref{fig:qualitygate}, and \ref{fig:qualitygatf}, nodes that were previously in areas of heavy congestion but now lie on roads experiencing light traffic, as the morning rush hour passes, are selected. Finally, the vehicle returns to the depot as its capacity constraints prevent it from serving additional customers, depicted in Figure \ref{fig:qualitygatg}. 
This example highlights the effectiveness of the SED2AM policy model in identifying and adapting to time-varying traffic conditions and making informed routing decisions based on the time-dependent travel times between nodes through its simultaneous encoder and decoder.

\FloatBarrier
\section{Conclusion} \label{section:conclusion}

In this study, we proposed a DRL-based method, named SED2AM, for effectively solving the MTTDVRP with maximum working hours constraints. The method integrates a Transformer-style policy model and utilizes a policy gradient-based DRL algorithm for training this policy network. SED2AM is distinguished by its simultaneous encoder, which captures dynamicity of travel times of vehicles, and a dual-decoding module. This decoding module includes a vehicle selection decoder for trip construction in multi-trip routing scenarios, and a trip construction decoder that constructs routes based on time-dependent and vehicle-specific contextual information in the state representation to ensure effective routing under maximum working hours constraints.

The experimental results on real-world instances from two major cities in Canada, Edmonton and Calgary, demonstrate that SED2AM not only outperforms heuristic-based and state-of-the-art DRL-based methods in terms of total travel time, but also operates much faster than heuristic-based solvers, while generalizing well to larger-scale problems. Finally, through conducting ablation studies, we investigated the impact of each of our major contributions.

The aim of this research is to take a step toward utilizing DRL-based methods for solving a more realistic and practical variant of VRP. Following this mindset, there are avenues for future research that we intend to investigate. First, we will investigate solving variants of the MTTDVRP that involve mixed pickup and delivery, as some real-world logistics applications, such as bottled beverage logistics, may involve a mixture of delivery and pickup to satisfy customers' needs. Second, we aim to consider time-window constraints for customers' demand, as in some applications, such as food delivery, deliveries to each customer should be made within a certain time window. 
In addition to incorporating these considerations on top of the existing considerations and constraints in the MTTDVRP with maximum working hours, we aim to further enhance the evaluation section of our work by performing a comprehensive comparison with two important studies: Zhao et al. \cite{zhao2024hybrid} and Pan et al. \cite{pan2021multi}, which propose innovative methods for solving the MTTDVRP. Furthermore, in our future work, we plan to perform more comprehensive experiments by investigating the impact of different RL algorithms on improving the policy model's routing performance. Finally, despite efforts to improve computational time—such as introducing temporal locality inductive bias to the simultaneous encoder and using a low-parameter network for vehicle selection—the increased complexity of the overall network resulted in higher computational times compared to DRL-based baselines. We plan to explore the adoption of a simpler policy model to improve the running time in the future. These improvements may include integrating sparse factorizations \cite{child2019generating} to manage computational demands efficiently, applying kernel-based linear transformations \cite{katharopoulos2020transformers} to reduce the computational complexity of the encoder's attention mechanism, and employing low-rank approximations \cite{wang2020linformer} to optimize the encoder's attention-based model for enhanced computational time.
\section{Acknowledgments}

The authors wish to extend their sincere gratitude to Dr. Yani Ioannou for his invaluable insights and the many productive discussions that greatly enriched this work. Aditionally, we thank Maryam Rezaie for her contributions in enhancing the visualizations presented. This project was supported in part by collaborative research funding from the National Research Council of Canada’s Artificial Intelligence for Logistics Program. The authors also would like to thank the Natural Sciences and Engineering Research Council of Canada (NSERC) for a part of financial support. Finally, we would like to express our sincere gratitude to the reviewer for their constructive feedback, which significantly improved the quality of this work.

\bibliographystyle{ACM-Reference-Format}
\bibliography{main}

\appendix

\section{Appendix A}
\label{appendixa}

The detailed summary of notations used throughout this paper is listed in Table \ref{tab:notations}.

\begin{footnotesize}
\begin{longtable}{|p{0.14\textwidth}|p{0.34\textwidth}|p{0.14\textwidth}|p{0.34\textwidth}|}
\caption{Summary of Notations} \label{tab:notations} \\
\hline
\textbf{Notation} & \textbf{Explanation} & \textbf{Notation} & \textbf{Explanation} \\
\hline
\endfirsthead

\multicolumn{4}{c}{{\bfseries Table \thetable\ continued from previous page}} \\
\hline
\textbf{Notation} & \textbf{Explanation} & \textbf{Notation} & \textbf{Explanation} \\
\hline
\endhead

\hline
\multicolumn{4}{r}{{Continued on next page}} \\
\hline
\endfoot

\hline
\endlastfoot
\( V=  \{v_0, \dots,  v_{n}\} \)           &   Node set   & \( E = \{e_{i,j} | 0 \leq i, j \leq n, i \neq j\} \)     & Edge set       \\ 
\hline
\( V_c \)         & Customer nodes set           & \( v_0 \)     & Depot node   \\ 
\hline
\( x_{v_i}, y_{v_i} \)           & Coordinates of the node $v_i$     & \( d_{v_i} \)     & Demand for node $v_i$        \\ 
\hline
\( TI \)         & Time-intervals set           & \( c(v_{i},v_{j},p) \)     & Travel time from $v_i$ to $v_j$ at time-interval $p$   \\ 
\hline
\( F \)           & Vehicles fleet set      & \( Q \)     & Maximum capacity        \\ 
\hline
\( T_{max} \)         & Vehicles' maximum working hours           & \( \mathcal{X} = \{\chi_{v_i}| v_i \in V\} \)     & Node-based feature vector   \\ 
\hline
\( \mathcal{E} =  \{\varepsilon_{ij,p}|0 \leq i, j \leq n, p \in TI\}\)           & Edge-based feature tensor      & \( S \)     &    State space   \\ 
\hline
\( A \)         & Action space           & \( \mathcal{T} \)     & Transition function   \\ 
\hline
\( r(s^{t},a^{t}) \)         & Reward for taking action $a^{t}$ in state $s^{t}$          & \(s_{F}^t =\left\{s^{t}_{1}, \dots, s^{t}_{f}\right\}\)     & Fleet state at time step t  \\ 
\hline
\( V^t_{visited} \)         & Set of nodes served up to time step $t$          & \( V^t_{to \textunderscore visit} \)     & Set of unserved customers up to time step $t$   \\ 
\hline
\( s_{R}^t \)         & Routing state at time step $t$           & \(  rc_i^{t} \)     & Vehicle $i$'s remaining capacity at time step $t$    \\ 
\hline
\( \mathcal{V}_i^{t} \)         & Current location of vehicle $i$           & \( \tau_i^{t} \)     & Vehicle $i$'s remaining working hours   \\ 
\hline
\( p_i^{t} \)         & Vehicle $i$'s current time-interval           & \( rt_i^{t} \)     & Time left in interval $p_i^{t}$ for vehicle $i$ \\ 
\hline
\hline
\( \pi_\theta(a^t|s^t) \)         & Stochastic policy parameterized by trainable parameter $\theta$           & \( p_\theta(\Pi|s_0) \)     & Conditional probability of solution $\Pi$ given the initial state $s_0$   \\ 
\hline
\( W \)         & Trainable weight matrix           & \( b \)     & Trainable bias   \\ 
\hline
\( h_{v_i,p}^{l} \)         & Time-dependent embedding of node \(v_i\) for interval \(p\) within encoder layer \(l\)           & \( h_{e_{ij,p}}^{l} \)     & Time-dependent embedding of edge \(e_{ij}\) for interval \(p\) within encoder layer \(l\)   \\ 
\hline
\( h_{v_i,p} = h^L_{v_i,p} \)         & Time-dependent node embedding for node \(v_i\) at interval \(p\) after \(L\) layers of encoding          & \( BN \)     & Batch normalized   \\ 
\hline
\( q^{l,m}_{i,p} \)         & Query for attention calculation of node \(v_i\) at interval \(p\), within layer \(l\), head \(m\)           & \( k^{l,m}_{i,p} \)     & Key for attention calculation of node \(v_i\) at interval \(p\), within layer \(l\), head \(m\)    \\ 
\hline
\( \nu^{l,m}_{i,p} \) &    Value for attention calculation of node \(v_i\) at interval \(p\), within layer \(l\), head \(m\)      & \( \epsilon^{l,m}_{ij,p} \)     & Linear projected edge $e_{ij}$'s embedding at interval \(p\), within layer \(l\), for head \(m\) \\
\hline
\( u^{l,m}_{ij,p} \)         & Compatibility of nodes \(v_i\) and \(v_j\) at interval \(p\), within layer \(l\), head \(m\) & \( d_k \)     & Key's embedding dimension   \\ 
\hline
\( \sigma \)        & Sigmoid activation function & \( a^{l,m}_{ij,p} \)     & Attention weight of node \(v_j\) for node \(v_i\) at interval \(p\), within layer \(l\), head \(m\)    \\ 
\hline
\( MHA^l_{i,p} \)         &  Multi-head attention operation & \( FF \)     & Feed-forward layer   \\ 
\hline
\( \omega^t_{vehicles} \)         & Vehicle status feature vector & \( e^t_{vehicles} \)     & Vehicle status embedding   \\ 
\hline
\( \omega^t_{td \textunderscore visited} \)         & Visited customers' time-dependent feature vector     & \( \Ddot{h}^i_{td} \) & Node embedding for the \(i\)-th customer at the respective time-interval   \\ 
\hline
\( e^t_{td \textunderscore visited} \)         & Visited customers' time-dependent embedding & \( \omega^{t,p}_{td \textunderscore to \textunderscore visit} \)     & To-be-visited customers' time-dependent feature vector   \\ 
\hline
\( e^t_{td \textunderscore to \textunderscore visit} \)         & To-be visited customers' time-dependent embedding & \( \omega^t_{indep \textunderscore  to \textunderscore visit} \)     & To-be-visited customers' time-independent feature vector   \\ 
\hline
\( e^t_{indep \textunderscore  to \textunderscore visit } \)     & To-be visited customers' time-independent embedding & \( p^{t,i}_{vehicles} \)     & Probability of choosing vehicle \(i\) for route expansion at the \(t\)-th decoding step   \\ 
\hline
\( e_{depot} \)         & Depot's embedding & \( softmax \)     & Softmax function   \\ 
\hline
\( h^t_{(c)} \)         & Context node embedding & \( {h^\prime}^t_{(c)} \)     & Node embeddings glimpse   \\ 
\hline
\( {u^\prime}^t_{(c),i} \)         & Context node embedding & \( \mathcal{M}^{t}_{v_i} \)     & Masking matrix element for node $v_i$ at the decoding step $t$   \\ 
\hline
\( C \)         &  Constant for controlling the entropy & \( p^{t,i}_{nodes} \)     & Probability of the node $v_i$ being selected at decoding step $t$   \\ 
\hline
\( \mathcal{L} \)         & Loss function & \( b(\Pi) \)     & Baseline estimator of loss   \\ 
\hline
\( F(\Pi|S) \)         & Fitness function for genetics algorithm & & \\
\hline
\end{longtable}
\end{footnotesize}

\end{document}